%% file: 0-main.tex
\RequirePackage{fix-cm}
\documentclass{article}
\PassOptionsToPackage{numbers, compress}{natbib}

\usepackage[main, final]{neurips_2026}

\usepackage[utf8]{inputenc} 
\usepackage[colorlinks=true]{hyperref}       
\hypersetup{
    pdftitle={LoCo: Selective Local Competition for Discriminative Open-Vocabulary Multi-Label Recognition},
    pdfauthor={Zhuoming Li, Bo Han, Xiaoyu Wang, Yuheng Jia},
    citecolor=blue,          
    linkcolor=black,         
    urlcolor=blue,           
}
\usepackage{booktabs}       
\usepackage{amsfonts}       
\usepackage{nicefrac}       
\usepackage{microtype}      
\usepackage{xcolor}         

\usepackage{enumitem} 
\setlist[enumerate]{left=0em, itemsep=0.2em, topsep=0.2em}

\usepackage{graphicx}
\usepackage{placeins}
\usepackage{multirow}
\usepackage{subcaption}
\DeclareMathSizes{8}{8}{5.5}{4.5}
\newcommand{\tablefont}{\fontsize{8}{9}\selectfont}
\usepackage{threeparttable}

\usepackage{xspace}
\newcommand{\eg}{\emph{e.g.}\xspace}
\newcommand{\ie}{\emph{i.e.}\xspace}

\usepackage{array}
\usepackage{colortbl}  
\usepackage{nicematrix}
\usepackage{tikz}
\usepackage{cleveref}

\title{LoCo: Selective Local Competition for Discriminative Open-Vocabulary Multi-Label Recognition}

\author{
  Zhuoming Li\textsuperscript{1} \quad Bo Han\textsuperscript{2} \quad Xiaoyu Wang\textsuperscript{1} \quad Yuheng Jia\textsuperscript{2,3}\thanks{Corresponding author.} \\
  {\small\normalfont \textsuperscript{1}School of Software Engineering, Southeast University, Nanjing, China} \\
  {\small\normalfont \textsuperscript{2}School of Computer Science and Engineering, Southeast University, Nanjing, China} \\
  {\small\normalfont \textsuperscript{3}Key Laboratory of New Generation Artificial Intelligence Technology and Its} \\
  {\small\normalfont Interdisciplinary Applications (Southeast University), Ministry of Education, China} \\
  {\small\normalfont\texttt{\char123 leezhuoming,hanbo,wang\_xy,yhjia\char125 @seu.edu.cn}}
}

\begin{document}

\maketitle

\begin{abstract}
  \input{1-abstract.tex}
\end{abstract}

\section{Introduction}
  \label{sec:introduction}
  \input{2-introduction.tex}

\section{Related Work}
  \label{sec:related_work}
  \input{3-literature.tex}

\section{Analysis: What Makes Local Prediction Work?}
  \label{sec:analysis}
  \input{4-analysis.tex}

\section{Method: LoCo Framework for Multi-label Recognition}
  \label{sec:method}
  \input{5-method.tex}

\section{Experiments}
  \label{sec:experiments}
  \input{6-experiments.tex}

\section{Conclusion}
  \label{sec:conclusion}
  \input{7-conclusion.tex}

\clearpage

\begin{ack}
This work was supported by the National Natural Science Foundation of China under Grants U24A20322 and 62576094. This research work was also supported by the Big Data Computing Center of Southeast University. The authors declare no competing interests.
\end{ack}

\bibliographystyle{unsrtnat}
\bibliography{9-references}


\appendix

\clearpage
\newpage

\input{8-appendix.tex}



\end{document}

%% file: 1-abstract.tex
Open-Vocabulary Multi-Label Recognition (OV-MLR) aims to identify all queried
semantic concepts present in an image. Existing methods mainly improve global
image-text matching or text-side representations, leaving local prediction
largely under-explored. In this paper, we revisit local prediction and reveal
that it can serve as a strong discriminative branch once two properties are
properly handled: spatial consistency of local visual features and
compatibility-aware local competition. This perspective is motivated by local
semantic sparsity: each image region is related to only a small subset of
queried concepts, while compatible concepts of different granularities or types
may still coexist locally.
We instantiate this perspective with \textbf{LoCo}, a selective \textbf{Lo}cal \textbf{Co}mpetition
framework for OV-MLR. LoCo adopts a spatially consistent visual encoder to
reduce local feature entanglement, and introduces Selective Softmax to impose
competition only among locally incompatible categories while preserving
compatible responses. 
Multi-scale inference further improves local prediction at different scales.
LoCo can also be ensembled with an existing global predictor to provide
complementary image-level context. 
Experiments on four multi-label benchmarks show that standalone LoCo improves
average F1-score and mAP over TagCLIP by \textbf{3.4\%} and \textbf{3.8\%}, respectively.
LoCo + DualPrompt achieves the highest averages among the compared methods, with gains of \textbf{4.8\%} and \textbf{5.2\%} over TagCLIP.
Our results suggest that local semantic sparsity offers a useful basis for
developing more discriminative open-vocabulary multi-label recognition methods.
The code is available at \url{https://github.com/lee-plus-plus/LoCo}.

%% file: 2-introduction.tex
Vision-language models, \eg, CLIP~\cite{CLIP}, have significantly advanced open-vocabulary visual recognition by aligning images and texts in a shared semantic space, where categories can be dynamically specified by natural language. Although CLIP enables single-label open-vocabulary recognition \cite{CuPL, CoOp, CaSED}, real-world visual understanding is inherently multi-label~\cite{emerging}: an image often contains multiple objects, scenes, and attributes simultaneously. Open-Vocabulary Multi-Label Recognition (OV-MLR) thus requires a model to recognize all semantic concepts present in an image, given arbitrary category names or textual descriptions, without supervised retraining \cite{TagCLIP}.

\begin{figure}[t]
    \centering
    \begin{subfigure}[b]{\linewidth}
        \centering
        \includegraphics[width=0.95\linewidth]{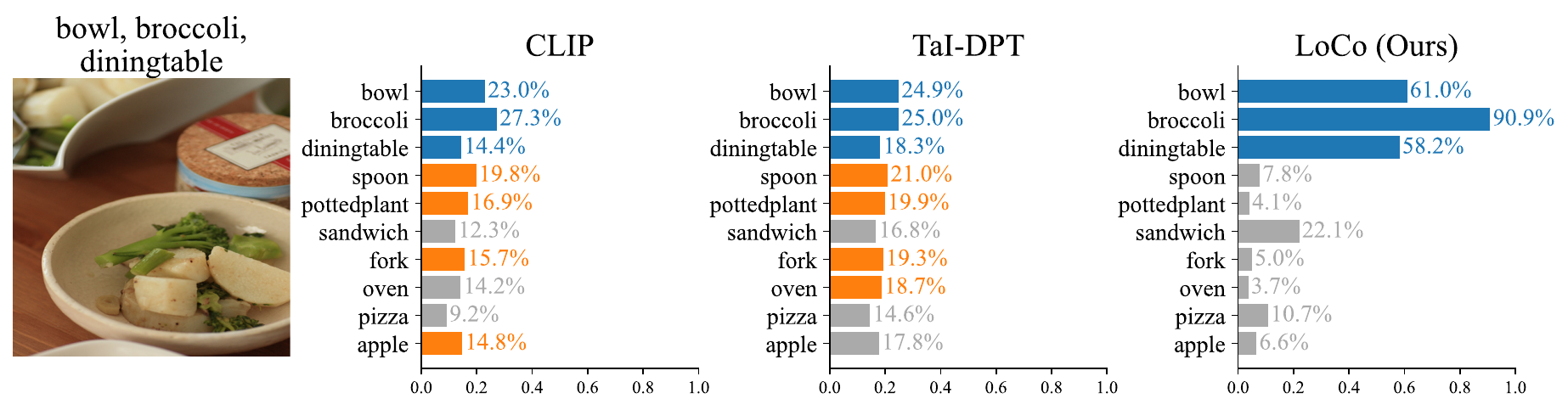}
        \subcaption{Existing OV-MLR methods CLIP~\cite{CLIP} and TaI-DPT~\cite{TaI-DPT} produce predictions with low discriminability.}
        \label{fig:low_discriminability}
    \end{subfigure}
    \begin{subfigure}[t]{0.60\linewidth}
        \centering
        \includegraphics[width=\linewidth]{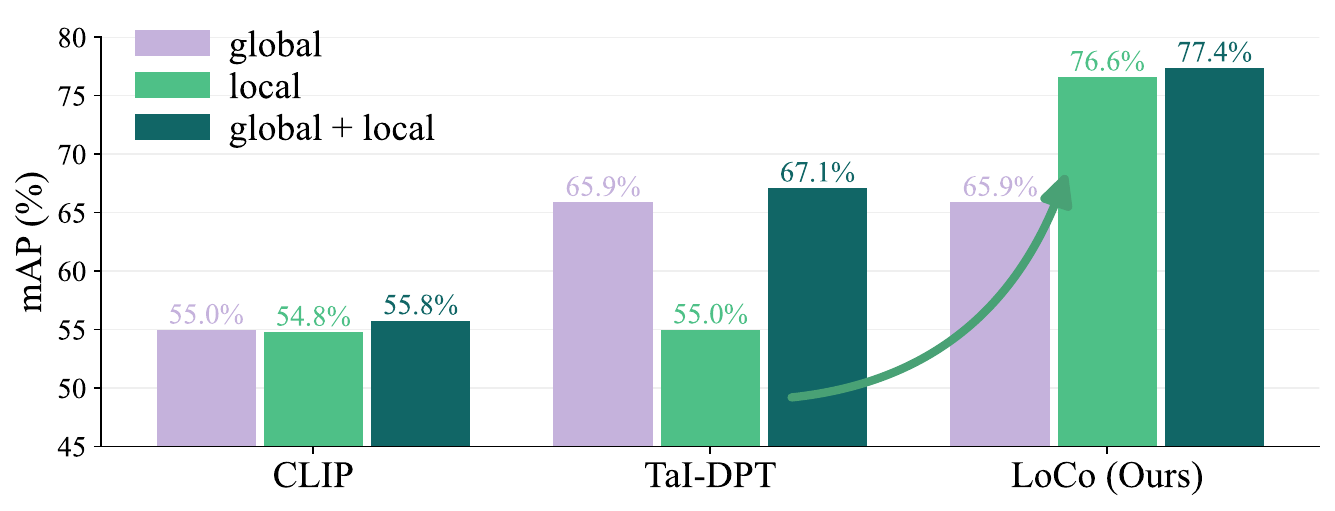}
        \subcaption{Our model (LoCo) unleashes the potential of local prediction.}
        \label{fig:local_potential}
    \end{subfigure}
    \begin{subfigure}[t]{0.39\linewidth}
        \centering
        \includegraphics[width=\linewidth]{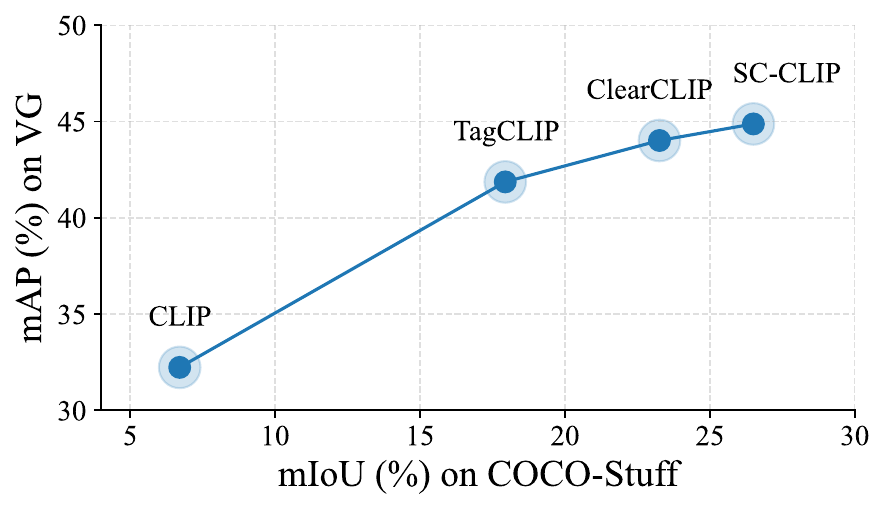}
        \subcaption{The segmentation capability of the visual backbone enables multi-label local prediction.}
        \label{fig:segmentation_enables_classification}
    \end{subfigure}
    \caption{Motivation and key insight of this paper. (a) Existing OV-MLR methods~\cite{CLIP, TaI-DPT} produce multi-label predictions with low discriminability, and our model overcomes this limitation. Blue bars denote true positive predictions, while orange bars denote false positive predictions. (b) The potential of local prediction is under-exploited by existing methods~\cite{CLIP, TaI-DPT}. Our model unleashes its potential and achieves outstanding performance. (c) The spatial consistency of the dense visual encoder, measured by its segmentation performance, \ie, mIoU, enables multi-label recognition in a local prediction framework.}
\end{figure}

Although previous OV-MLR studies have made progress \cite{DualCoOp, TaI-DPT, TaI++, CoMC, Hier-TaI, T2I-PAL, ConCEPT, DualPrompt}, achieving discriminative multi-label predictions remains challenging. Existing methods produce prediction scores with a relatively small margin between positive and negative categories, making it difficult to obtain discriminative predictions, as shown in \Cref{fig:low_discriminability}.
This phenomenon reflects two key challenges in OV-MLR: (1) \textbf{visual feature entanglement}. Multiple semantic concepts are compressed into a single visual representation, which dilutes category-specific cues~\cite{BiAM}. (2) \textbf{weak response separability}. Text features encoded by CLIP are highly correlated~\cite{Unmix-CLIP} (shown in Appendix~\ref{sec:score_separability}). Since CLIP-based category scores are computed from cosine similarities between visual features and textual queries, class-independent predictions tend to produce small numerical gaps among category scores.
Existing OV-MLR methods mainly improve prediction scores from the text side~\cite{TaI-DPT,TaI-Adapter,TaI++,Unmix-CLIP,T2I-PAL,ConCEPT}, which partially alleviates the weak response separability challenge, leaving the feature entanglement challenge relatively underexplored. 

To achieve discriminative multi-label prediction, we argue that \textbf{local prediction} has the potential to address both of the aforementioned challenges. 
Unlike global features that include multiple coexisting semantics, local features have much sparser semantic compositions, a property we term \textbf{local semantic sparsity}, which makes local prediction different from global multi-label prediction. 
Previous methods~\cite{TaI-DPT,TaI-Adapter,CoMC,DualPrompt} usually implement local prediction as a spatial repetition of image-level prediction and use it only as a supplementary branch, leaving much of its potential under-exploited, as shown in \Cref{fig:local_potential}. This also suggests that once the bottlenecks of visual feature entanglement and weak response separability are alleviated, local prediction can bring substantial improvements over existing state-of-the-art text-side methods.

Based on this observation, we investigate how to unleash the potential of local prediction under the local semantic sparsity prior.
First, to satisfy the local semantic sparsity prior, local visual features should be spatially consistent. CLIP's~\cite{CLIP} raw local visual features also suffer from feature entanglement, resulting in noisy spatial responses~\cite{TagCLIP,Unmix-CLIP,SCLIP} and thus unreliable local prediction. \Cref{fig:segmentation_enables_classification} provides empirical evidence that stronger spatial consistency in spatial responses (mIoU) is beneficial for local multi-label recognition (mAP).
Second, to exploit local semantic sparsity, a suitable inter-class competition mechanism should be introduced. Inter-class competition can effectively enlarge the relative gap between positive and negative category prediction scores in single-label prediction, but it may incorrectly suppress compatible categories in multi-label scenarios.

To this end, we propose a selective \textbf{Lo}cal \textbf{Co}mpetition framework (\textbf{LoCo}) for OV-MLR. LoCo first leverages a spatially consistent visual encoder, SC-CLIP~\cite{SC-CLIP}, to provide more reliable local visual features. It then designs Selective Softmax, which imposes class competition on local regions while considering both exclusiveness and compatibility between categories, thereby improving the discriminability of multi-label scores. In addition, LoCo incorporates multi-scale inference using grid cropping to enhance the detection of small objects and local details, and can be integrated with other global prediction methods.
The main contributions of this work are summarized as follows:

\begin{enumerate}
    \item We identify spatial consistency of local visual features as a key prerequisite for effective local prediction in OV-MLR, and empirically show that spatially consistent local features can turn local prediction into a strong discriminative branch.

    \item We propose Selective Softmax, a local competition mechanism that introduces competition between each category and its locally incompatible competitors. It improves response discriminability while avoiding the over-suppression of compatible concepts.

    \item We develop LoCo, an open-vocabulary multi-label recognition framework that improves average F1-score and mAP over the most competitive baseline, TagCLIP~\cite{TagCLIP}, by 3.4\% and 3.8\%, respectively. LoCo also complements existing global predictors: combining it with DualPrompt's global predictions~\cite{DualPrompt} further improves average F1-score and mAP by 1.4\% each over standalone LoCo.
\end{enumerate}

%% file: 3-literature.tex
Multi-label learning \cite{emerging} is a fundamental yet challenging task in computer vision, which requires models to identify multiple semantic concepts present in an image. Considering the high cost of collecting multi-label annotations, previous literature has extensively focused on weakly supervised settings \cite{DiCaP, PML-CD, NLR}. In recent years, Open-Vocabulary Multi-Label Recognition (OV-MLR) has been proposed to leverage the open-category and zero-shot capability of vision-language models \cite{CLIP} for multi-label learning.
Early studies mainly focused on adapting CLIP from single-label recognition \cite{CoOp} to multi-label scenarios \cite{DualCoOp, CDUL}. 
Most existing OV-MLR methods focus on improving the quality of textual queries. For example, Text-as-Image~\cite{TaI-DPT, TaI-Adapter, CoMC, TaI++} methods convert text into proxy visual samples to obtain class prototypes that are more suitable for multi-label prediction via prompt tuning. Later works \cite{RCNn, TaI-HnGL, T2I-PAL, ConCEPT, Hier-TaI, DualPrompt} further exploit external language knowledge, including semantic granularity, contextual descriptions, category relations, and co-occurrence knowledge. 
Score calibration and compositional prompts were also discussed \cite{SPARC}.

Apart from the aforementioned methods that mainly focus on the text side, another line of work exploits local visual information. Early OV-MLR methods \cite{DualCoOp,TaI-DPT} combine global and local responses as dual-granularity prediction, which has since been widely adopted by later works \cite{CoMC, TaI++, T2I-PAL, DualPrompt, Unmix-CLIP}. These works demonstrate the value of local information, but still treat local prediction as a spatial repetition of image-level prediction. TagCLIP~\cite{TagCLIP} developed a local-to-global framework that directly uses CLIP local features for open-vocabulary multi-label prediction. 

A closely related area is training-free Open-Vocabulary Segmentation (OVS), which systematically studies how to obtain spatially consistent dense visual representations from CLIP~\cite{CLIP-ES,MaskCLIP,CLIP-Surgery,AutoSeg}. 
SCLIP~\cite{SCLIP} replaces final-layer query--key attention with query--query and key--key correlations, while ClearCLIP~\cite{ClearCLIP} uses self-self attention and removes the final residual connection and feed-forward network (FFN). SC-CLIP~\cite{SC-CLIP} uses intermediate-feature similarity for self-calibration and multi-level feature fusion. 
Other approaches~\cite{DINOiser, ProxyCLIP, Talk2DINO} transfer spatial information from self-supervised visual models~\cite{DINOv2} to dense vision-language prediction.

\begin{figure}[t]
    \centering
    \includegraphics[width=\linewidth]{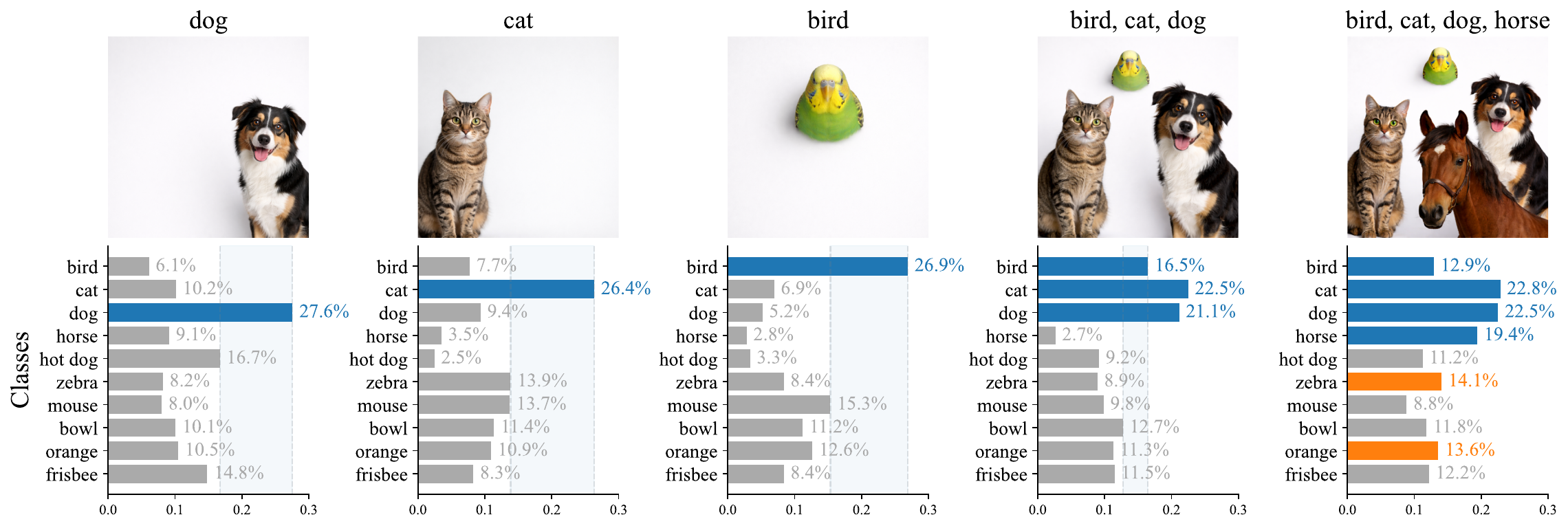}
    \caption{Feature entanglement in multi-label scenarios. Blue bars denote true positive predictions, while orange bars denote false positive predictions. As more semantic concepts co-occur in the same image, CLIP's prediction score gap between positive and negative classes narrows. We attribute this to feature entanglement caused by mutual dilution across different categories of information.}
    \label{fig:feature_entanglement}
\end{figure}

%% file: 4-analysis.tex
\begin{figure}[t]
    \centering
    \subcaptionbox{t-SNE of the dense feature map. \label{fig:backbone_tsne_featuremap}}{%
        \includegraphics[height=6.3cm]{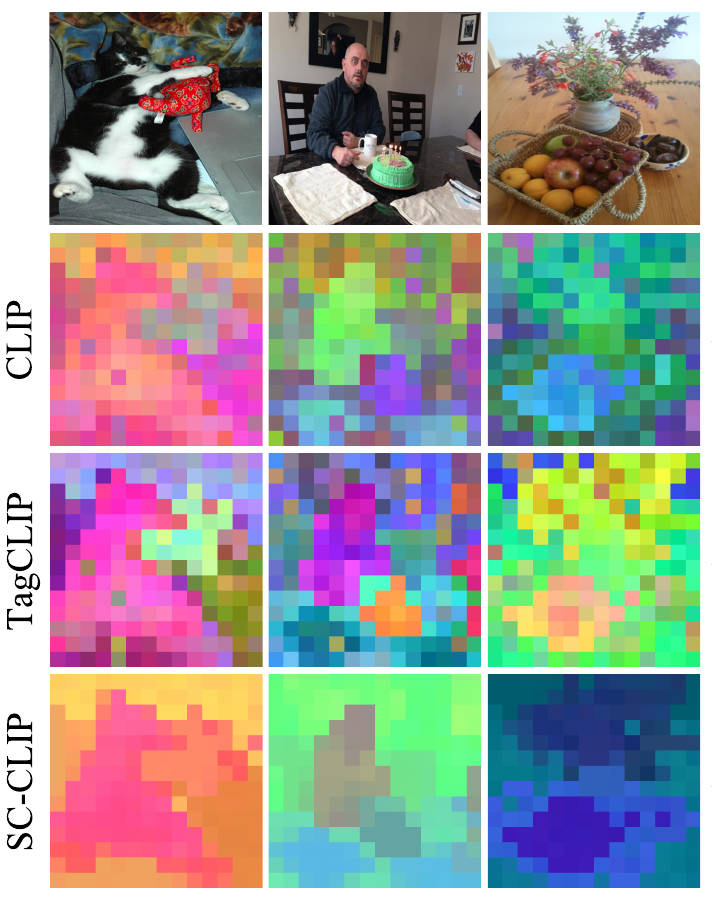}%
    }%
    \hfill
    \subcaptionbox{Dense predictions for different classes. \label{fig:backbone_dense_prediction}}{%
        \includegraphics[height=6.3cm]{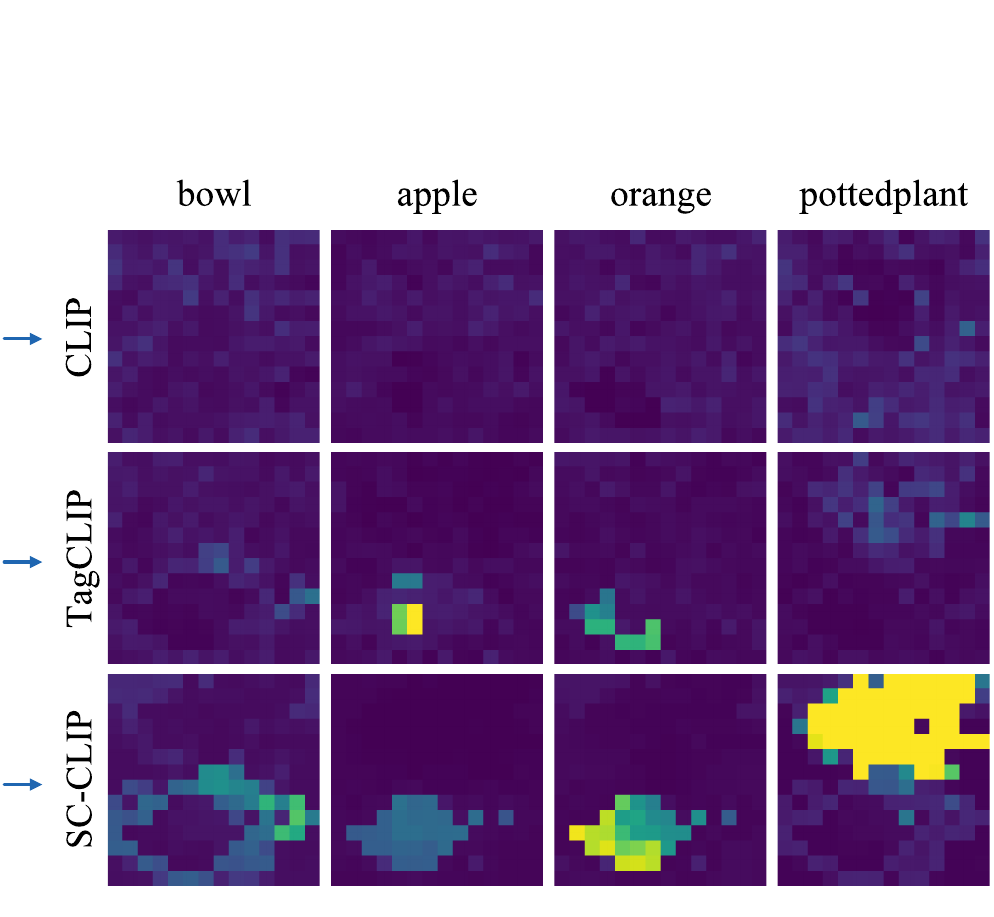}%
    }%
    \hfill
    \subcaptionbox{Local scores. \label{fig:backbone_local_scores}}{%
        \includegraphics[height=6.3cm]{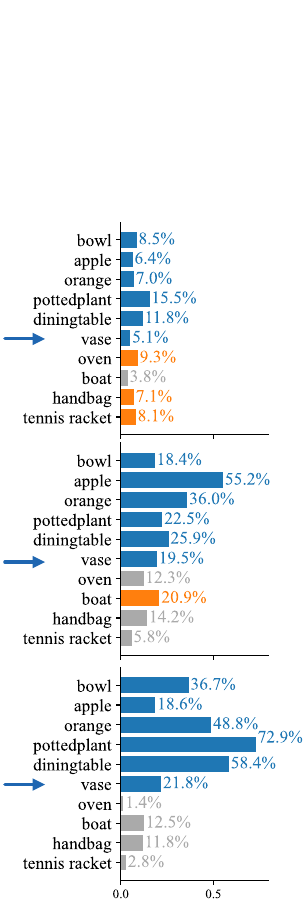}%
    }%
    \caption{Spatial consistency of dense visual backbones matters for discriminative local prediction. CLIP~\cite{CLIP}, TagCLIP~\cite{TagCLIP}, and SC-CLIP~\cite{SC-CLIP} are selected as weak, moderate, and strong visual backbones. Softmax is applied at each spatial position to produce dense predictions. With stronger visual backbones, local features are less fragmented and more spatially consistent, thereby producing more discriminative multi-label predictions.}
    \label{fig:backbone_analysis}
\end{figure}

\subsection{Local Prediction Remains Auxiliary}
\label{sec:analysis_local_prediction_auxiliary}

\Cref{fig:feature_entanglement} illustrates a typical challenge in multi-label prediction: feature entanglement on global features weakens the discriminability of global prediction. This phenomenon motivates the widely used global-local framework in OV-MLR \cite{DualCoOp, TaI-DPT}. Since local features preserve fine-grained visual evidence and avoid global aggregation, they are expected to alleviate feature entanglement and produce more discriminative category responses.
However, \Cref{fig:local_potential} shows that prediction based on CLIP's local features remains weak. Moreover, combining local predictions with global predictions brings only limited improvement, which suggests that local prediction in existing frameworks remains auxiliary, rather than serving as a reliable branch. We identify that this is mainly because local features still suffer from entanglement, as discussed below.

\subsection{Local Features Lack Spatial Consistency}
\label{sec:analysis_local_dense_encoder}

Effective local prediction is based on the prerequisite that local visual features are disentangled, \ie, they reflect the semantics of their corresponding image regions instead of mixing information from unrelated regions. However, this does not naturally hold for CLIP's default visual encoder, which is trained with image-text contrastive supervision without explicit spatial consistency regularization. \Cref{fig:backbone_tsne_featuremap} shows that CLIP's local features are fragmented, spatially inconsistent, and semantically mixed. Therefore, \textbf{local features are also entangled}, which results in unreliable local activation and thus non-discriminative local prediction, as shown in \Cref{fig:backbone_dense_prediction} and \Cref{fig:backbone_local_scores}. This observation suggests that the bottleneck lies in the spatial consistency of local visual features.

To quantify the spatial consistency of local features, we adopt the semantic segmentation metric mIoU as an external proxy, which is appropriate because semantic segmentation directly evaluates whether local features can produce spatially coherent and boundary-aware category responses, matching the aforementioned requirement.
Therefore, rather than designing a new spatial encoder specifically for OV-MLR, we leverage recent progress in open-vocabulary segmentation, \eg, ClearCLIP~\cite{ClearCLIP} and SC-CLIP~\cite{SC-CLIP}, to obtain more spatially consistent local features.

We do not claim the dense visual backbone itself as our contribution. Instead, we use it to verify an important finding: local prediction becomes effective only when local representations are spatially consistent. \Cref{fig:segmentation_enables_classification} suggests a correlation between segmentation capability and MLR performance, and \Cref{fig:backbone_analysis} provides explainable evidence for its necessity.
These facts allow us to decouple the design of dense feature representations from OV-MLR, and instead focus on how reliable local features can be effectively used for discriminative multi-label prediction in the following analysis.

\subsection{From Local Over-Competition to Selective Competition}
\label{sec:analysis_selective_competition}

\begin{table}[t]
    \centering
    \setlength{\tabcolsep}{4pt}
    \caption{Major types of semantic relations among categories in the VG dataset.}
    \label{tab:category_compatibility_types}
    {
        \begin{threeparttable}
            \setlength{\tabcolsep}{4pt}
            \tablefont
            \begin{tabular}{p{0.29\linewidth} p{0.40\linewidth} p{\dimexpr0.31\linewidth-24pt\relax}}
                \toprule
                \textbf{Relation type} & \textbf{Examples}                         & \textbf{Locally Compatible?} \\
                \midrule

                Scene-level Co-occurrence
                                       & \emph{room--bed}; \emph{street--car}      & No                           \\
                Covering Relation
                                       & \emph{person--shirt}; \emph{hand--glove}  & Usually Yes                  \\
                Overlap Relation
                                       & \emph{snow--mountain}; \emph{light--lamp} & Usually Yes                  \\
                Taxonomic Inclusion
                                       & \emph{animal--dog}; \emph{vehicle--car}   & Yes                          \\
                Part--whole Composition
                                       & \emph{person--arm}; \emph{car--wheel}     & Yes                          \\
                Surface Marking
                                       & \emph{sign--letter}; \emph{screen--logo}  & Yes                          \\
                Near-synonymy
                                       & \emph{airplane--plane}; \emph{cap--hat}   & Yes                          \\
                \bottomrule
            \end{tabular}
        \end{threeparttable}
    }
\end{table}

\begin{figure}[t]
    \centering
    \includegraphics[width=\linewidth]{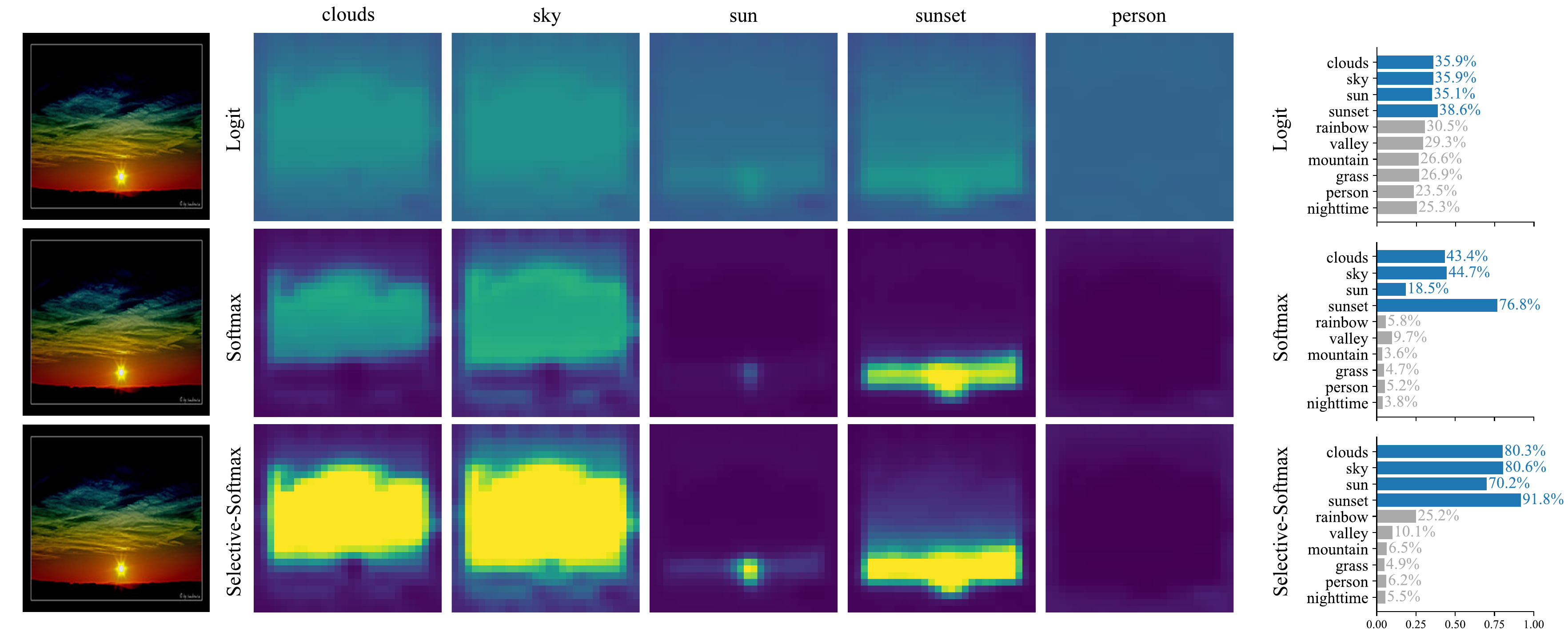}
    \caption{Multi-label local prediction with different prediction strategies. Softmax is effective for enhancing discriminability, but it introduces unreasonable competition between compatible classes, \eg, \textit{clouds} and \textit{sky}, or \textit{sun} and \textit{sunset}, resulting in over-competition. Selective Softmax introduces competition between each category and its locally incompatible competitors and produces better predictions.}
    \label{fig:over_competition}
\end{figure}

When each image region reflects its own semantics, spatially consistent local features enable local category competition in practice. Applying softmax over local category responses can enlarge relative score differences and produce more discriminative predictions, which is widely adopted in semantic segmentation.
However, this strategy implicitly assumes local semantic uniqueness, \ie, each local region should correspond to only one semantic category. While this assumption is natural in semantic segmentation, where each pixel is treated as a single-label prediction, it is too restrictive for OV-MLR, whose predefined category set may contain concepts of different granularities and types. For example, queried categories may include coarse and fine categories (\textit{animal} and \textit{dog}), or whole and part (\textit{person} and \textit{arm}). \Cref{tab:category_compatibility_types} summarizes the major types of semantic relationships among categories in the VG dataset, where many categories can coexist in the same region.
Therefore, treating these categories as mutually exclusive would incorrectly suppress their dense predictions and limit the model's capability to handle open-vocabulary semantics. We refer to this property as \textbf{local semantic sparsity}, which assumes that each region can belong to a few concepts and is a better prior than local semantic uniqueness for OV-MLR.

To exploit local semantic sparsity, inter-class competition should be introduced selectively, \ie, between each category and its locally incompatible competitors, avoiding direct competition between compatible categories. \Cref{fig:over_competition} shows that raw logit scores are under-competitive, softmaxed scores are over-competitive, and selectively softmaxed scores provide structured local competition that better fits the semantic relationships in OV-MLR. Implementation details of Selective Softmax are described in \Cref{sec:selective_competition}.

%% file: 5-method.tex
\begin{figure}[t]
    \centering
    \includegraphics[width=\linewidth]{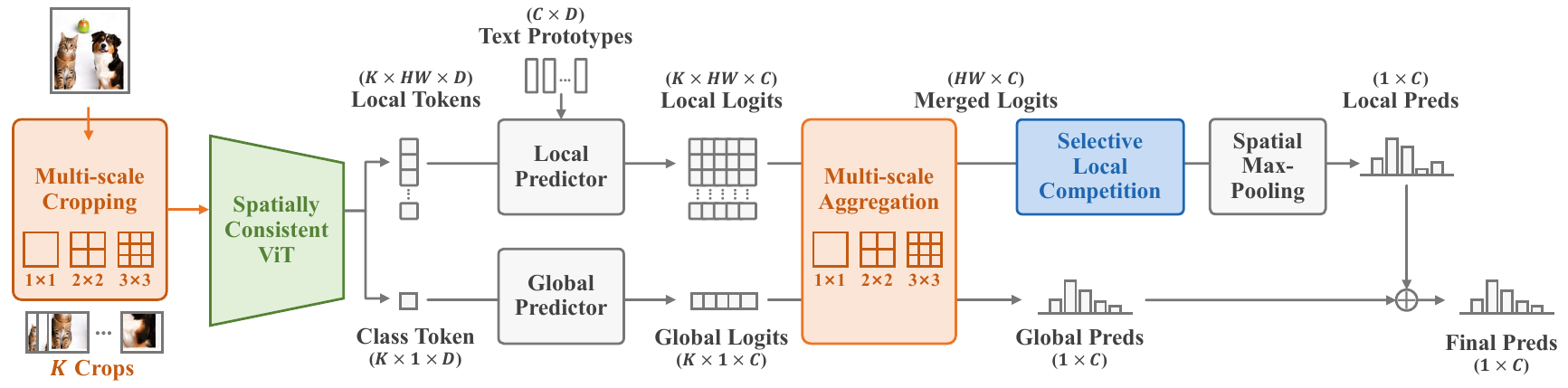}
    \caption{Framework of the proposed LoCo. 
    An image is first divided into \(K\) crops using grid cropping and encoded by a spatially consistent ViT. Local tokens produce crop-level logits. The multi-scale aggregation module spatially aligns and averages these logits before Selective Softmax and spatial max pooling. The resulting local predictions can optionally be ensembled with global predictions averaged over the same crops. }
    \label{fig:model}
\end{figure}

\Cref{sec:analysis} has already shown that effective local prediction for
open-vocabulary multi-label recognition requires two conditions: spatially consistent local features and a discriminative prediction mechanism that considers local exclusiveness and compatibility between categories. To achieve them, we propose
\textbf{LoCo}, a selective local competition framework for open-vocabulary multi-label recognition, as shown in \Cref{fig:model}. 
It adopts a \textbf{spatially consistent dense visual encoder} (\Cref{sec:spatial_backbone}) to produce local tokens for image-level local prediction. \textbf{Selective local competition} (\Cref{sec:selective_competition}) enhances local prediction, while \textbf{multi-scale inference} (\Cref{sec:multi_scale_aggregation}) captures objects at different scales. Scores from an existing global predictor can optionally be ensembled with LoCo's local scores.

\subsection{Preliminaries}
\label{sec:preliminaries}

We first revisit the default CLIP pipeline for multi-label recognition. Given an image
\(x\), the ViT-based visual encoder \(E_\mathrm{v}\) divides it into \(H \times W\)
patches and outputs a sequence of visual tokens
\begin{equation}
    [\mathbf{v}_{\mathrm{cls}}, \mathbf{v}_1,\mathbf{v}_2,\ldots,\mathbf{v}_{HW}]
    = E_\mathrm{v}(x),
\end{equation}
where \(\mathbf{v}_{\mathrm{cls}}\) is the global \texttt{[CLS]} token and
\(\mathbf{v}_i\) denotes the \(i\)-th local patch token. Given a category name set \(\mathcal{S}=\{s_1,s_2,\ldots,s_C\}\) for classes \(1, 2, \ldots, C\), each category name is converted
into a textual prompt $\mathrm{Prompt}(\cdot)$ and encoded by the CLIP text encoder \(E_\mathrm{t}\) as
\begin{equation}
    \mathbf{q}_c = E_\mathrm{t}(\mathrm{Prompt}(s_c)), \quad c=1,2,\ldots,C .
\end{equation}
The default CLIP pipeline uses only the global token to compute the
image-level multi-label prediction score via cosine similarity
\begin{equation}
    \hat{\mathbf{y}}^\mathrm{g} = [
        \langle \mathbf{v}_{\mathrm{cls}}, \mathbf{q}_1 \rangle,
        \langle \mathbf{v}_{\mathrm{cls}}, \mathbf{q}_2 \rangle,
        \ldots,
        \langle \mathbf{v}_{\mathrm{cls}}, \mathbf{q}_C \rangle
    ].
\end{equation}

\subsection{Spatially Consistent Visual Encoder}
\label{sec:spatial_backbone}


Following the analysis in \Cref{sec:analysis_local_dense_encoder}, we adopt an existing spatially consistent encoder instead of training a new one. CLIP's last ViT block is a major source of local spatial inconsistency: its query--key attention mixes information across regions, and the final residual connection and FFN further propagate global context into patch tokens. 
Existing training-free corrections modify these operations to improve spatial consistency, \eg, 
SCLIP~\cite{SCLIP} replaces the final-layer query--key attention with query--query and key--key correlations while keeping the residual and FFN paths, whereas ClearCLIP~\cite{ClearCLIP} uses self-self attention and drops the final residual connection and FFN.

In our implementation, we adopt SC-CLIP~\cite{SC-CLIP} as LoCo's spatially consistent encoder. Specifically, SC-CLIP applies intermediate-layer similarity for self-calibration and multi-level feature fusion. For efficiency, we disable only its anomaly-token resolving module, which refines segmentation boundaries but introduces extra inference cost. 
All encoder operations are inherited from SC-CLIP and introduce no training or learnable parameters, and LoCo applies Selective Softmax to the resulting dense logits.
Formally, the spatially consistent visual backbone \(E_\mathrm{v}'\) outputs
\begin{equation}
    [\mathbf{v}_{\mathrm{cls}}, \mathbf{v}_1',\mathbf{v}_2',\ldots,\mathbf{v}_{HW}']
    = E_\mathrm{v}'(x),
\end{equation}
where \(\mathbf{v}_i'\) denotes the \(i\)-th spatially consistent local patch token.

\subsection{Selective Local Competition}
\label{sec:selective_competition}

\begin{figure}[t]
    \centering
    \includegraphics[width=\linewidth]{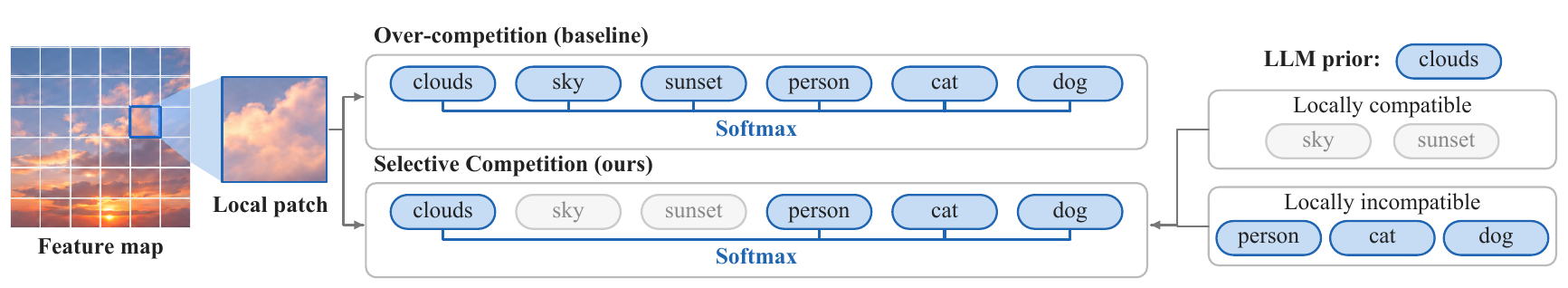}
    \caption{Selective local competition guided by an LLM compatibility prior, where \textit{clouds} is selected as the target category.}
    \label{fig:selective_competition_illustration}
\end{figure}

Given spatially consistent local features, each local region produces a
category logit vector through vision-language matching. Let
\begin{equation}
    \mathbf{s}_i = [
        \langle \mathbf{v}_i', \mathbf{q}_1 \rangle,
        \langle \mathbf{v}_i', \mathbf{q}_2 \rangle,
        \ldots,
        \langle \mathbf{v}_i', \mathbf{q}_C \rangle
    ]
\end{equation}
denote the local logits on the \(i\)-th local token. 
A class-independent prediction treats all
categories separately and therefore lacks relative competition, while a
standard softmax over all categories assumes full mutual exclusiveness and
may incorrectly suppress compatible concepts. To address this issue, we
introduce selective local competition, where each category only competes with
categories that are incompatible with it, as shown in \Cref{fig:selective_competition_illustration}.

In open-vocabulary multi-label recognition, the category set may contain concepts of different granularities and
abstraction levels. Ignoring such concept relations can either place
compatible concepts into unnecessary competition or leave mutually exclusive
concepts completely independent. For example, \textit{sky} and \textit{cloud} can coexist as
a background region and its visible component. Rather than explicitly
distinguishing all possible relation types, we simplify these relationships to \textbf{local
compatibility}. Concepts that can jointly describe the same local region are treated as compatible, while the others are regarded as competitors.

To obtain such compatibility knowledge, we use an LLM to perform explicit
semantic reasoning over the category name set, whose details are shown in \Cref{sec:compatibility_prior_analysis}.
Although CLIP is effective at measuring the similarity between an image region and a text query, it does not directly provide semantic compatibility or incompatibility relations among category names. 
In contrast, LLMs encode richer conceptual knowledge and are better suited for reasoning over category relations. 
Therefore, we feed the complete category name set \(\mathcal{S}=\{s_1,s_2,\ldots,s_C\}\) into the LLM with a structured prompt, asking which categories can coexist with each category within the same local region. 

Specifically, we parse the structured response offline and apply symmetrization: two categories are connected whenever either directed compatibility list includes the other. The resulting local compatibility graph is reused for all images with that category vocabulary. 
Its neighbors define the compatible set \(\mathcal{A}^{+}_c \subseteq \{1,\ldots,C\}\) for each category \(c\). We then define the competition set of \(s_c\) as all categories except itself and its compatible categories
\begin{equation}
    \mathcal{A}^{-}_c
    =
    \{j \mid j\neq c,\ j\notin \mathcal{A}^{+}_c\}.
\end{equation}
Selective Softmax recalibrates
the local logits by applying softmax to each category together with its incompatible competitors. For category \(c\) and the \(i\)-th token, its selective competition score is
\begin{equation}
    \hat{y}^\mathrm{l}_{i,c}
    = \mathrm{SelSoftmax}_c\left( \mathbf{s}_i \right)
    = \frac
    {\exp\left( \mathbf{s}_{i,c} / \tau \right)}
    {\exp\left( \mathbf{s}_{i,c} / \tau \right)+ \sum_{c' \in \mathcal{A}^{-}_{c}} \exp\left(\mathbf{s}_{i,c'} / \tau \right)},
\end{equation}
where \(\tau\) is a temperature parameter. 
The score \(\hat{y}^\mathrm{l}_{i,c}\) measures the relative advantage of category \(s_c\)
relative to its incompatible competitors. Thus, Selective Softmax introduces
competition where mutual exclusiveness is expected, while preserving
compatibility among concepts.

\subsection{Multi-scale Inference}
\label{sec:multi_scale_aggregation}

LoCo performs multi-scale inference using grid cropping with \(1\times1\), \(2\times2\), and \(3\times3\) partitions, yielding \(K=14\) image crops. Each crop is
resized to \(224\times224\) pixels and processed by the dense visual encoder
to obtain a local logit map
\(\mathbf{s}_w\in\mathbb{R}^{H\times W\times C}\).
Each map is then bilinearly interpolated to the corresponding crop region
on a common \(H\times W\) grid spanning the original image and positioned
according to the crop offset. The aligned logits are averaged at each
grid location:
\begin{equation}
    \bar{s}_{u,c}
    =
    \frac{1}{|\mathcal{W}(u)|}
    \sum_{w\in\mathcal{W}(u)}
    s^{\mathrm{aligned}}_{w,u,c},
\end{equation}
where \(\mathcal{W}(u)\) denotes the set of crops covering location \(u\).
This produces a merged logit map
\(\bar{\mathbf{s}}\in\mathbb{R}^{H\times W\times C}\).
We then apply Selective Softmax to the merged logits, followed by spatial
max pooling to obtain image-level local predictions:
\begin{equation}
    \hat{y}^{\mathrm{l}}_c
    =
    \max_u \mathrm{SelSoftmax}_c
    \left(\bar{\mathbf{s}}_u\right),
\end{equation}
where \(\bar{\mathbf{s}}_u=[\bar{s}_{u,1},\ldots,\bar{s}_{u,C}]\).
While local scores capture region-level
semantics, labels such as scenes, attributes, and contextual concepts still
benefit from image-level representations.
LoCo can optionally be combined with an
existing global predictor \(g\). Applying
\(g\) to all \(K\) crops yields one global score vector per crop,
\(\hat{\mathbf{y}}^{\mathrm{g}}_w=g(x_w,\mathcal{S})\), and the image-level
global score is their unweighted mean:
\begin{equation}
    \hat{\mathbf{y}}^{\mathrm{g}}
    =
    \frac{1}{K}
    \sum_{w=1}^{K}
    \hat{\mathbf{y}}^{\mathrm{g}}_w .
\end{equation}
The final prediction is
obtained by
\begin{equation}
    \hat{\mathbf{y}}
    =
    \alpha \hat{\mathbf{y}}^\mathrm{g}
    +
    (1-\alpha) \hat{\mathbf{y}}^\mathrm{l},
\end{equation}
where \(\alpha\) weights the global prediction.
LoCo thus serves as a local branch to which existing global OV-MLR predictors can be attached.

%% file: 6-experiments.tex
\newcommand{\best}[1]{\textbf{\underline{#1}}}
\newcommand{\secondbest}[1]{\underline{#1}}
\newcommand{\good}[1]{\secondbest{#1}}
\newcommand{\pz}{\phantom{0}}

\newlength{\threecharwidth}
\settowidth{\threecharwidth}{COCO}
\newcolumntype{C}{>{\centering\arraybackslash}p{\threecharwidth}}
\newcolumntype{A}{>{\columncolor{gray!10}\centering\arraybackslash}p{\threecharwidth}}
\newcommand{\tightmidrule}{%
    \specialrule{\lightrulewidth}{0pt}{0pt}%
}
\newcommand{\padbeforemidrule}{%
    \rule[-\dimexpr\dp\strutbox+\aboverulesep\relax]{0pt}
    {\dimexpr\ht\strutbox+\dp\strutbox+\aboverulesep\relax}%
}

\newcommand{\padaftermidrule}{%
    \rule{0pt}{\dimexpr\ht\strutbox+\belowrulesep\relax}%
}

\subsection{Experimental Setup}

\paragraph{Datasets.}
\label{sec:datasets}
We evaluate our method on four widely used multi-label recognition datasets: Pascal VOC 2007 (VOC), MS-COCO 2014 (COCO), NUS-WIDE (NUS), and Visual Genome 150 (VG). Performance on the test sets is reported. VOC \cite{VOC2007} and COCO \cite{COCO2014} have segmentation-level annotations and thus exclusive categories, whereas NUS \cite{NUS-WIDE} and VG \cite{VG, VG150} contain more complex non-exclusive categories, which are shown in \Cref{tab:category_compatibility_types}.
These benchmarks cover 20--150 categories and different image-level label densities.

\paragraph{Baselines.}
We compare LoCo with state-of-the-art OV-MLR methods: the original CLIP~\cite{CLIP}; text-prompt- or text-prototype-based methods including TaI-DPT~\cite{TaI-DPT}, TaI-Adapter~\cite{TaI-Adapter}, CoMC~\cite{CoMC}, and DualPrompt~\cite{DualPrompt}; the local-prediction-based method TagCLIP~\cite{TagCLIP}; and the composition-based method SPARC~\cite{SPARC}.

\paragraph{Evaluation metrics.}
We use mean average precision (mAP) and F1-score, macro-averaged across categories, as the main evaluation metrics.
\textbf{mAP} measures the ranking quality of category predictions, which is widely adopted in previous literature~\cite{TaI-DPT,T2I-PAL,ConCEPT,TagCLIP}.
\textbf{F1-score} evaluates the binarized prediction after thresholding, which quantifies the practical discriminability of multi-label prediction.
For each method and dataset, we select the threshold maximizing validation F1-score over 0.01--0.99 (step 0.01) and apply it to the test set. 
Component ablations also report \textbf{Recall@10}, the fraction of annotated positives retrieved among each image's 10 highest-scoring categories, and micro \textbf{PR-AUC}, computed over image--category pairs within each dataset, to fully evaluate their capability.

\paragraph{Implementation details.}
For fair comparison, all methods adopt ViT-B/16 as the visual backbone and are evaluated with the same category sets. For text-as-image methods, the same LLM-generated caption datasets are used for text-as-image training. Each caption dataset has 40,000 captions and 2\(\sim\)5 labels per sample. LoCo uses a Selective Softmax temperature of $\tau = 1/40$, multi-scale inference using grid cropping with \(1\times1\), \(2\times2\), and \(3\times3\) partitions, and \(\alpha = 1/3\) for optional global ensembling. Unless otherwise specified, results are obtained under the same input resolution of \(224\times224\) and the same text prompt templates. As an exception, TagCLIP uses an input resolution of \(448\times448\) and introduces additional background categories following its original setting. All experiments are conducted on a server equipped with NVIDIA GeForce RTX 3090 \(\times\) 8 GPUs and an AMD EPYC 7K62 48-Core Processor.

\subsection{Comparison Results}

\Cref{tab:main_results} compares LoCo with state-of-the-art methods on four benchmark datasets. Standalone LoCo achieves the highest average F1-score and mAP among the standalone methods, with average gains of 3.4\% in F1-score and 3.8\% in mAP over TagCLIP.
When ensembled with global predictions from TaI-DPT, LoCo achieves 59.7\% F1-score and 66.4\% mAP. LoCo + DualPrompt achieves the highest averages of 59.9\% F1-score and 67.4\% mAP, with gains of 4.8\% and 5.2\% over TagCLIP.
The consistent gains across diverse datasets validate the proposed local competition mechanism and spatially consistent local representations in enhancing multi-label discriminability.

\begin{table}[t]
    \centering
    \setlength{\tabcolsep}{3pt}
    \caption{Experimental results on open-vocabulary multi-label recognition. mAP measures test-set ranking quality. \textit{Avg.} is the unweighted mean over four datasets. \best{Best} and \good{second-best} results are highlighted.}
    \label{tab:main_results}
    {
        \begin{threeparttable}
            \csname TPT@hookin\endcsname{NiceTabular}
            \tablefont
            \settowidth{\threecharwidth}{COCO}
            \begin{NiceTabular}{l l CCCCC CCCCC}
                \CodeBefore
                    \tikz \fill[gray!10]
                        ([yshift=-\belowrulesep]2-|7) rectangle (13-|8)
                        ([yshift=-\belowrulesep]2-|12) rectangle (13-|13);
                \Body
                \toprule
                \multirow{2}{*}{Method}    & \multirow{2}{*}{Venue}
                                           & \multicolumn{5}{c}{F1-score (\%)}
                                           & \multicolumn{5}{c}{mAP (\%)}                                                                                                                                                                        \\
                \cmidrule(lr){3-7} \cmidrule(lr){8-12}
                                           &                                   & VOC         & COCO        & NUS         & VG          & \textit{Avg.} & VOC         & COCO        & NUS         & VG          & \textit{Avg.}\padbeforemidrule \\
                \tightmidrule \padaftermidrule
                CLIP                       & ICML'21                           & 65.2 & 39.0 & 30.8 & 21.1 & 39.0 & 79.1        & 55.0        & 40.6        & 31.2        & 51.5                            \\
                TaI-DPT\tnote{*}                    & CVPR'23                           & 75.4 & 55.2 & 41.8 & 37.6 & 52.5 & 86.1        & 66.0        & 46.2        & 41.2        & 59.9                            \\
                TaI-Adapter\tnote{*}                & arXiv'23                          & 77.6 & 55.1 & 31.7 & 33.3 & 49.4 & 87.1        & 61.3        & 45.0        & 35.5        & 57.2                            \\
                CoMC\tnote{*}                       & ICML'24                           & 76.8 & 55.5 & 34.2 & 37.2 & 50.9 & 88.3        & 66.8        & 47.1        & 38.6        & 60.2                            \\
                TagCLIP                    & AAAI'24                           & \best{85.3} & 61.6 & 35.8 & 37.8 & 55.1 & \good{92.5}        & 72.3        & 41.7        & 42.5        & 62.2                            \\
                SPARC                      & CVPR'25                           & 72.7 & 53.9 & 33.5 & 34.0 & 48.5 & 87.6        & 66.2        & 46.5        & 39.0        & 59.8                            \\
                DualPrompt\tnote{\dag}                 & ICLR'26                           & 76.0 & 57.8 & 38.4 & 37.2 & 52.4 & 87.8        & 69.5        & \good{48.1} & 36.5        & 60.5\padbeforemidrule \\
                \tightmidrule \padaftermidrule
                LoCo                       & Ours                              & 83.1 & 65.1 & 40.7 & 45.0 & 58.5 & \best{93.0}        & 76.6        & 45.3        & 49.2        & 66.0                            \\
                LoCo + TaI-DPT\tnote{*}             & Ours                              & \good{84.2} & \good{65.7} & \good{43.0} & \good{45.8} & \good{59.7} & \good{92.5}        & \good{77.4} & 46.4        & \good{49.3} & \good{66.4}                     \\
                LoCo + DualPrompt\tnote{\dag}          & Ours                              & 84.1 & \best{66.3} & \best{43.3} & \best{45.9} & \best{59.9} & \best{93.0} & \best{78.3} & \best{48.2} & \best{49.9} & \best{67.4}                     \\
                \bottomrule
            \end{NiceTabular}
            \begin{tablenotes}
                \item[*] Extra training required. \item[\dag] Extra annotations required.
            \end{tablenotes}
        \end{threeparttable}
    }
\end{table}

\subsection{Ablation Study}

\begin{table}[t]
    \centering
    \begin{minipage}[t]{0.46\linewidth}
        \vspace{0pt}
        \centering
        \captionsetup{type=table}
        \caption{Ablation of the standalone local branch. \textit{Sel.} denotes Selective Softmax and \textit{M.S.} denotes 14-crop multi-scale inference. \best{Best} and \good{second-best} are highlighted.}
        \label{tab:ablation}
        \tablefont
        \setlength{\tabcolsep}{2pt}
        \begin{tabular}{>{\raggedright\arraybackslash}p{36pt}>{\centering\arraybackslash}p{15pt}>{\centering\arraybackslash}p{15pt}*{4}{>{\centering\arraybackslash}p{22pt}}}
            \toprule
            \multirow{2}{*}[-6pt]{Backbone} & \multirow{2}{*}[-6pt]{\textit{Sel.}} & \multirow{2}{*}[-6pt]{\textit{M.S.}}
             & \multicolumn{4}{c}{\textit{Avg.} (\%)} \\
            \cmidrule(lr){4-7}
             & & & \shortstack{F1-\\score} & \raisebox{4pt}[0pt][0pt]{\parbox[c][2\baselineskip][c]{\linewidth}{\centering mAP}} & \shortstack{Recall\\@10} & \shortstack{PR-\\AUC}\padbeforemidrule \\
            \tightmidrule \padaftermidrule
            CLIP &  &  & 46.4 & 55.0 & 71.6 & 43.1 \\
            ClearCLIP &  &  & 52.5 & 59.8 & 79.8 & 51.0 \\
            ClearCLIP & \checkmark &  & 53.8 & 60.0 & 81.8 & 54.0 \\
            ClearCLIP &  & \checkmark & 54.7 & 63.7 & 81.6 & 53.7 \\
            ClearCLIP & \checkmark & \checkmark & 56.0 & 64.0 & 83.5 & 56.8 \\
            SC-CLIP &  &  & 54.9 & 62.4 & 82.2 & 54.7 \\
            SC-CLIP & \checkmark &  & \good{56.7} & 62.9 & 83.5 & \good{57.8} \\
            SC-CLIP &  & \checkmark & 56.6 & \good{65.5} & \good{83.9} & 56.6 \\
            SC-CLIP & \checkmark & \checkmark & \best{58.5} & \best{66.0} & \best{85.1} & \best{59.8} \\
            \bottomrule
        \end{tabular}
    \end{minipage}
    \hfill
    \begin{minipage}[t]{0.52\linewidth}
        \vspace{0pt}
        \centering
        \captionsetup{type=table}
        \caption{Fair comparison between LoCo and baseline method TaI-DPT for local prediction with the same dense encoder (Backbone) and the same multi-scale inference (\textit{M.S.}).}
        \label{tab:matched_summary}
        \tablefont
        \setlength{\tabcolsep}{2pt}
        \begin{tabular}{>{\raggedright\arraybackslash}p{32pt}>{\raggedright\arraybackslash}p{31pt}>{\centering\arraybackslash}p{18pt}*{4}{>{\centering\arraybackslash}p{23pt}}}
            \toprule
            \multirow{2}{*}[-6pt]{Backbone} & \multirow{2}{*}[-6pt]{Method} & \multirow{2}{*}[-6pt]{\textit{M.S.}}
             & \multicolumn{4}{c}{\textit{Avg.} (\%)} \\
            \cmidrule(lr){4-7}
             & & & \shortstack{F1-\\score} & \raisebox{4pt}[0pt][0pt]{\parbox[c][2\baselineskip][c]{\linewidth}{\centering mAP}} & \shortstack{Recall\\@10} & \shortstack{PR-\\AUC}\padbeforemidrule \\
            \tightmidrule \padaftermidrule
            CLIP & TaI-DPT & & 52.5 & 59.4 & 75.2 & 47.1 \\
            CLIP & TaI-DPT & \checkmark & 51.1 & 60.2 & 75.6 & 47.9 \\
            SC-CLIP & TaI-DPT & & 55.8 & 63.1 & 81.1 & 54.6 \\
            SC-CLIP & TaI-DPT & \checkmark & 55.7 & \good{64.8} & 81.1 & 54.7 \\
            CLIP & LoCo & & 45.9 & 52.0 & 72.4 & 43.6 \\
            CLIP & LoCo & \checkmark & 47.8 & 55.3 & 74.4 & 47.0 \\
            SC-CLIP & LoCo & & \good{56.7} & 62.9 & \good{83.5} & \good{57.8} \\
            SC-CLIP & LoCo & \checkmark & \best{58.5} & \best{66.0} & \best{85.1} & \best{59.8} \\
            \bottomrule
        \end{tabular}
    \end{minipage}
\end{table}

\begin{table}[t]
    \centering
\begin{minipage}[t]{0.40\linewidth}
        \vspace{0pt}
        \centering
    \setlength{\tabcolsep}{4pt}
        \captionsetup{type=table}
        \caption{Comparison of different compatibility priors. GPT-5.5 Prior is finally adopted.}
        \label{tab:compatibility_ablation}
        \setlength{\tabcolsep}{2pt}
            \tablefont
            \begin{tabular}{lcc}
                \toprule
                Rule / Prior & F1-score & mAP \\
                \midrule
                Raw logits    & 53.1        & 62.8 \\
                Softmax       & 56.6        & \good{65.5} \\
                \midrule
                Random Prior  & 52.9        & 61.5 \\
                WordNet Prior       & 57.0        & \good{65.5} \\
                GPT-5.5 Prior       & \good{58.5} & \best{66.0} \\
                LLMs Consensus Prior & \best{58.6} & \best{66.0} \\
                \bottomrule
            \end{tabular}
    \end{minipage}
    \hfill
    \begin{minipage}[t]{0.58\linewidth}
        \vspace{0pt}
\centering
        \captionsetup{type=table}
        \caption{Computation cost and efficiency of standalone methods. Size is the image's width and height; crop counts are specified for LoCo.}
        \label{tab:computation_resources}
        \setlength{\tabcolsep}{1.5pt}
            \tablefont
            \begin{tabular}{lc rrr}
                \toprule
                Method (size; crops) & mAP & FLOPs & Throughput & Peak Mem \\
                \midrule
                CLIP (224; --)        & 51.5 & 35.1 G/img  & 380.9 img/s & 0.8 GB \\
                TaI-DPT (224; --)     & 59.9 & 35.3 G/img  & 377.6 img/s & 0.8 GB \\
                TagCLIP (224; --)     & 61.0 & 70.6 G/img  & 25.1 img/s  & 1.2 GB \\
                TagCLIP (448; --)     & 62.2 & 228.4 G/img & 8.4 img/s   & 9.3 GB \\
                \midrule
                LoCo (224; 1)        & 62.9 & 35.2 G/img  & 342.1 img/s & 1.6 GB \\
                LoCo (224; 14)       & 66.0 & 493.4 G/img & 27.8 img/s  & 5.8 GB \\
                \bottomrule
            \end{tabular}
    \end{minipage}
\end{table}

\paragraph{Visual backbone matters.}
\Cref{tab:ablation} evaluates the contributions of LoCo's components.
Under Softmax and single-crop inference, replacing CLIP with ClearCLIP
and SC-CLIP improves average F1-score from 46.4\% to 52.5\% and 54.9\%,
respectively, and mAP from 55.0\% to 59.8\% and 62.4\%.
These improvements highlight the importance of spatially consistent
local features for local prediction.

\paragraph{Selective Softmax consistently contributes.}
\Cref{tab:ablation} shows that, across ClearCLIP and SC-CLIP with single- and multi-scale inference,
Selective Softmax improves average F1-score by 1.3--1.9\% and mAP
by 0.3--0.5\%. It also improves Recall@10 by 1.2--2.0\% and PR-AUC
by 3.0--3.2\%. These consistent gains across four evaluation metrics
support the effectiveness of selective local competition in improving
multi-label discriminability. Per-dataset results are provided
in \Cref{sec:cross_backbone_metrics}.

\paragraph{Multi-scale inference further helps.}
Across different visual backbones (ClearCLIP or SC-CLIP) and local competition conditions (with and without Selective Softmax), 
multi-scale inference improves average F1-score by 1.7--2.2\%, mAP by 3.1--4.0\%, Recall@10 by 1.6--1.8\%, and PR-AUC by 1.9--2.8\%, demonstrating the benefit of capturing local evidence at different scales.

\paragraph{LoCo better exploits spatial consistency and inference budget.}
\Cref{tab:matched_summary} compares LoCo with TaI-DPT under the same visual backbone and inference settings for local prediction. TaI-DPT benefits from both components: replacing CLIP with SC-CLIP increases average mAP from 59.4\% to 63.1\%, and multi-scale inference further raises it to 64.8\%. With SC-CLIP and multi-scale inference, LoCo outperforms TaI-DPT across all four metrics. These results show that LoCo makes more effective use of spatially consistent local features and multi-scale inference than TaI-DPT.

\subsection{Local Compatibility Prior Analysis}

\paragraph{Non-competition and over-competition harm.}
\Cref{tab:compatibility_ablation} examines the need for selective competition and the choice of compatibility prior under the same local model and 14-crop inference. Selective Softmax with the GPT-5.5 prior achieves 58.5\% F1-score and 66.0\% mAP, outperforming both raw logits without competition (53.1\% and 62.8\%) and full Softmax with excessive competition (56.6\% and 65.5\%). This comparison supports the necessity of selectively imposing competition among locally incompatible categories. 

\paragraph{LLM generates reliable category compatibility.}
\Cref{tab:compatibility_ablation} compares four compatibility prior strategies.
The random prior randomly assigns compatibility between category pairs,
while the WordNet prior derives compatibility from is-a, part-of, and
synonym relations. The GPT-5.5 prior uses a compatibility graph generated
by GPT-5.5, whereas the LLM consensus takes the intersection of
graphs generated independently by four different LLMs.
GPT-5.5 outperforms both random and WordNet priors in F1-score and mAP,
supporting LLM-based compatibility construction. The consensus yields
a marginally higher F1-score (58.6\% vs.\ 58.5\%) with the same mAP (66.0\%).
We adopt the single GPT-5.5 prior for simplicity, given its comparable
performance without combining multiple LLMs.
Details are provided in \Cref{sec:prior_stability,sec:wordnet_prior}.

\subsection{Computational Cost Analysis}
\label{sec:computation_resources}

\Cref{tab:computation_resources} shows that LoCo without \textit{M.S.} matches the original CLIP in FLOPs and throughput, whereas \textit{M.S.} inference raises its cost by about 14 times due to grid cropping, giving an \textbf{accuracy-efficiency trade-off}. Despite higher total FLOPs, LoCo with \textit{M.S.} still keeps a throughput advantage over TagCLIP at high resolution, because its crop-level computation is fully parallelizable while TagCLIP relies on a less hardware-friendly two-stage pipeline. The trade-off is acceptable, since high-resolution recognition is often necessary for small objects.


%% file: 7-conclusion.tex
This paper revisits the role of local prediction in OV-MLR. 
We show that local prediction becomes a strong discriminative branch when local visual features are spatially consistent and category competition is aware of local semantic compatibility. Motivated by the observation of local semantic sparsity, we propose LoCo, a selective local competition framework that uses spatially consistent dense features and Selective Softmax to impose
competition only among locally incompatible categories.
Experiments on four multi-label benchmarks show that standalone LoCo improves average F1-score and mAP over TagCLIP by 3.4\% and 3.8\%, respectively. With the optional DualPrompt global predictor, LoCo achieves the highest averages among the compared methods, improving over TagCLIP by 4.8\% in F1-score and 5.2\% in mAP.
Beyond this framework, our findings suggest that local semantic sparsity offers a useful basis for OV-MLR and motivate future work on local semantic structure and dense vision-language representations.

%% file: 8-appendix.tex
\section*{Table of Contents for the Appendix}
\makeatletter
\setcounter{tocdepth}{2}
\@starttoc{apx}
\let\appendixaddcontentsline\addcontentsline
\renewcommand{\addcontentsline}[3]{%
  \def\appendixcontentsfile{#1}\def\standardcontentsfile{toc}%
  \ifx\appendixcontentsfile\standardcontentsfile
    \appendixaddcontentsline{apx}{#2}{#3}%
  \else
    \appendixaddcontentsline{#1}{#2}{#3}%
  \fi
}
\makeatother

\FloatBarrier
\clearpage

\section{Text Correlation and Response Separability}
\label{sec:score_separability}

\Cref{tab:text_cosine_similarity} reports mean pairwise cosine similarity among category text prototypes for the four benchmark vocabularies. CLIP has higher mean pairwise similarity than TaI-DPT in each vocabulary. These statistics describe text-prototype geometry and complement the response-distribution analysis in \Cref{tab:response_separability}.

\begin{table}[!htbp]
    \centering
    \setlength{\tabcolsep}{4pt}
    \caption{Mean text-prototype cosine similarities, displayed in units of $10^{-2}$. Multiply displayed similarities by $10^{-2}$ to recover the dimensionless cosine values.}
    \label{tab:text_cosine_similarity}
    \tablefont
    \begin{tabular}{lcrr}
        \toprule
        Dataset & Classes & CLIP mean & TaI-DPT mean \\
        \midrule
        VOC & 20 & 81.0 & 24.0 \\
        COCO & 80 & 76.0 & 14.0 \\
        NUS & 81 & 80.0 & 19.0 \\
        VG & 150 & 78.0 & 17.0 \\
        \bottomrule
    \end{tabular}
\end{table}

\Cref{tab:response_separability} summarizes the reported response distributions across the four datasets, separating positive-label and negative-label scores. Raw scores use different scales across methods. Standardized separation is reported using Cohen's $D$:
\begin{equation}
    D = \frac{\mu^{+}-\mu^{-}}{s_{\mathrm{pool}}},
    \label{eq:cohen_d}
\end{equation}
where $\mu^{+}$ and $\mu^{-}$ are the positive and negative score means, and $s_{\mathrm{pool}}$ is the pooled estimated population standard deviation. The reported $D$ values are 1.5 for CLIP, 2.3 for TaI-DPT, and 4.1 for LoCo. These are descriptive comparisons of complete model predictions.

\begin{table}[!htbp]
    \centering
    \setlength{\tabcolsep}{4pt}
    \caption{Reported response statistics across four datasets. All score statistics are displayed in units of $10^{-3}$; multiply them by $10^{-3}$ to recover the original response scale. Cohen's $D$ is dimensionless and unscaled.}
    \label{tab:response_separability}
    \tablefont
    \begin{tabular}{lccr}
        \toprule
        \multirow{2}{*}{Model} & Positive only & Negative only & \multirow{2}{*}{Cohen's $D$} \\
        & Mean $\pm$ Std. & Mean $\pm$ Std. & \\
        \midrule
        CLIP & $175.0\pm32.0$ & $138.0\pm24.0$ & 1.5 \\
        TaI-DPT & $17.0\pm5.0$ & $11.0\pm3.0$ & 2.3 \\
        LoCo & $449.0\pm248.0$ & $55.0\pm78.0$ & 4.1 \\
        \bottomrule
    \end{tabular}
\end{table}

\FloatBarrier
\section{Supplementary Ablation Results}
\label{sec:supplementary_ablations}

\subsection{Per-Dataset Ablation Results}
\label{sec:cross_backbone_metrics}

\Cref{tab:ablation_details} compares Softmax and Selective Softmax across ClearCLIP and SC-CLIP with one or 14 crops. The table reports F1-score, mAP, micro PR-AUC, and Recall@10 (all in \%). Recall@10 is evaluated against annotated positive labels. Averages are unweighted means over the four datasets; \best{best} and \good{second-best} ranks use unrounded scores.

Selective Softmax improves the reported dataset-averaged metrics in all four backbone/crop configurations. The individual results include a slight decline in VG micro PR-AUC for 14-crop ClearCLIP: both values round to 40.5\%. This metric-specific exception is distinct from the VG F1-score improvements in \Cref{tab:ablation_details}.

\begin{table}[!htbp]
    \centering
    \setlength{\tabcolsep}{3.5pt}
    \caption{Per-dataset ablation of the standalone local branch. Top: F1-score and mAP. Bottom: micro PR-AUC and Recall@10. \textit{Sel.} denotes Selective Softmax and \textit{M.S.} denotes 14-crop inference; unchecked settings use Softmax and one crop, respectively.}
    \label{tab:ablation_details}
    {
        \tablefont
        \settowidth{\threecharwidth}{COCO}
            \begin{NiceTabular}{lcc CCCCC CCCCC}
                \CodeBefore
                    \tikz \fill[gray!10] ([yshift=-\belowrulesep]2-|8) rectangle (11-|9);
                    \tikz \fill[gray!10] ([yshift=-\belowrulesep]2-|13) rectangle (11-|14);
                    \tikz \fill[gray!10] ([yshift=-\belowrulesep]12-|8) rectangle (21-|9);
                    \tikz \fill[gray!10] ([yshift=-\belowrulesep]12-|13) rectangle (21-|14);
                \Body
                \toprule
                \multirow{2}{*}{Backbone}
                 & \multirow{2}{*}{\textit{Sel.}}
                 & \multirow{2}{*}{\textit{M.S.}}
                 & \multicolumn{5}{c}{F1-score (\%)}
                 & \multicolumn{5}{c}{mAP (\%)} \\
                \cmidrule(lr){4-8} \cmidrule(lr){9-13}
                 & & & VOC & COCO & NUS & VG & \textit{Avg.} & VOC & COCO & NUS & VG & \textit{Avg.}\padbeforemidrule \\
                \tightmidrule \padaftermidrule
                ClearCLIP &  &  & 76.2 & 60.7 & 34.8 & 38.2 & 52.5 & 88.3 & 70.8 & 37.9 & 42.0 & 59.8 \\
                ClearCLIP & \checkmark &  & 78.1 & 61.0 & 36.2 & 39.9 & 53.8 & 88.5 & 70.6 & 38.4 & 42.7 & 60.0 \\
                ClearCLIP &  & \checkmark & 77.9 & 65.0 & 35.2 & 40.8 & 54.7 & 91.2 & \best{77.1} & 40.2 & 46.4 & 63.7 \\
                ClearCLIP & \checkmark & \checkmark & 80.0 & \best{65.1} & 36.6 & 42.4 & 56.0 & 91.3 & \good{76.9} & 40.8 & 46.9 & 64.0 \\
                SC-CLIP &  &  & 79.6 & 60.7 & 39.2 & 40.0 & 54.9 & 90.8 & 71.1 & 43.1 & 44.4 & 62.4 \\
                SC-CLIP & \checkmark &  & \good{82.0} & 61.3 & \best{41.0} & 42.5 & \good{56.7} & 91.1 & 70.9 & 44.2 & 45.5 & 62.9 \\
                SC-CLIP &  & \checkmark & 79.9 & 64.7 & 39.0 & \good{42.6} & 56.6 & \good{92.8} & 76.8 & \good{44.3} & \good{48.2} & \good{65.5} \\
                SC-CLIP & \checkmark & \checkmark & \best{83.1} & \good{65.1} & \good{40.7} & \best{45.0} & \best{58.5} & \best{93.0} & 76.6 & \best{45.3} & \best{49.2} & \best{66.0} \\
                \midrule
                \multirow{2}{*}{Backbone}
                 & \multirow{2}{*}{\textit{Sel.}}
                 & \multirow{2}{*}{\textit{M.S.}}
                 & \multicolumn{5}{c}{PR-AUC (\%)}
                 & \multicolumn{5}{c}{Recall@10 (\%)} \\
                \cmidrule(lr){4-8} \cmidrule(lr){9-13}
                 & & & VOC & COCO & NUS & VG & \textit{Avg.} & VOC & COCO & NUS & VG & \textit{Avg.}\padbeforemidrule \\
                \tightmidrule \padaftermidrule
                ClearCLIP &  &  & 75.0 & 60.0 & 31.9 & 37.0 & 51.0 & 99.6 & 85.7 & 82.6 & 51.3 & 79.8 \\
                ClearCLIP & \checkmark &  & 79.7 & 63.0 & 36.1 & 37.2 & 54.0 & 99.7 & 89.5 & 85.6 & 52.3 & 81.8 \\
                ClearCLIP &  & \checkmark & 76.8 & 65.0 & 32.5 & 40.5 & 53.7 & \good{99.9} & 88.6 & 83.4 & 54.5 & 81.6 \\
                ClearCLIP & \checkmark & \checkmark & 81.6 & \good{67.8} & 37.3 & 40.5 & 56.8 & \good{99.9} & \best{92.3} & 86.3 & 55.4 & 83.5 \\
                SC-CLIP &  &  & 80.1 & 62.6 & 37.0 & 39.1 & 54.7 & 99.6 & 88.4 & 87.8 & 53.2 & 82.2 \\
                SC-CLIP & \checkmark &  & \good{84.1} & 65.8 & \best{41.5} & 39.9 & \good{57.8} & 99.6 & 89.6 & \good{89.9} & 55.0 & 83.5 \\
                SC-CLIP &  & \checkmark & 80.8 & 66.6 & 36.8 & \good{42.3} & 56.6 & \good{99.9} & 90.8 & 88.7 & \good{56.2} & \good{83.9} \\
                SC-CLIP & \checkmark & \checkmark & \best{84.9} & \best{69.7} & \good{41.4} & \best{43.0} & \best{59.8} & \best{99.9} & \good{92.1} & \best{90.6} & \best{57.9} & \best{85.1} \\
                \bottomrule
            \end{NiceTabular}
    }
\end{table}

\subsection{Temperature Sensitivity}
\label{sec:sensitivity_analysis}

\begin{figure}[h]
    \centering
    \includegraphics[width=\linewidth]{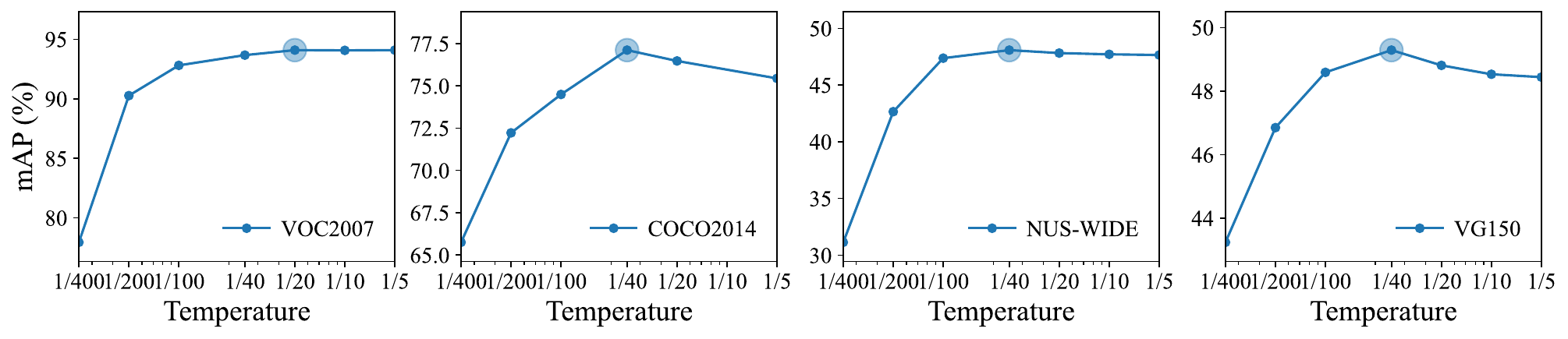}
    \caption{Sensitivity analysis of the hyperparameter \(\tau\). }
    \label{fig:sensitivity_tau}
\end{figure}

We analyze the sensitivity of the Selective Softmax temperature \(\tau\) in \Cref{fig:sensitivity_tau}. Overall, the best results are achieved at \(\tau=1/40\) on 3/4 benchmarks, while on the remaining dataset \(\tau=1/20\) is only slightly better than \(\tau=1/40\). This suggests that the proposed method is not highly sensitive to the exact temperature choice around this range. In particular, the mAP performance remains relatively stable near \(\tau=1/40\), indicating that LoCo is robust to moderate changes of the temperature parameter. Therefore, we use \(\tau=1/40\) as the default setting in all experiments.

\FloatBarrier
\section{Matched Encoder and Inference Comparisons}
\label{sec:matched_details}

\Cref{tab:matched_details} expands the matched comparison summarized in \Cref{tab:matched_summary} to per-dataset values. It compares LoCo with TaI-DPT using the same SC-CLIP dense encoder, with and without \textit{M.S.}, alongside the original CLIP results. All averages are unweighted means over the four datasets; scores are rounded to one decimal after calculation and highlighted ranks use unrounded values.

\begin{table}[!htbp]
    \centering
    \setlength{\tabcolsep}{4pt}
    \caption{Per-dataset matched comparison against TaI-DPT on the same SC-CLIP backbone, with and without \textit{M.S.} Top: F1-score and mAP. Bottom: Recall@10 and micro PR-AUC. Recall@10 is evaluated against annotated positive labels.}
    \label{tab:matched_details}
    {
        \tablefont
        \settowidth{\threecharwidth}{COCO}
            \begin{NiceTabular}{llc ccccC ccccC}
                \CodeBefore
                    \tikz \fill[gray!10] ([yshift=-\belowrulesep]2-|8) rectangle (9-|9);
                    \tikz \fill[gray!10] ([yshift=-\belowrulesep]2-|13) rectangle (9-|14);
                    \tikz \fill[gray!10] ([yshift=-\belowrulesep]10-|8) rectangle (17-|9);
                    \tikz \fill[gray!10] ([yshift=-\belowrulesep]10-|13) rectangle (17-|14);
                \Body
            \toprule
            \multirow{2}{*}{Backbone} & \multirow{2}{*}{Method} & \multirow{2}{*}{\textit{M.S.}} & \multicolumn{5}{c}{F1-score (\%)} & \multicolumn{5}{c}{mAP (\%)} \\
            \cmidrule(lr){4-8} \cmidrule(lr){9-13}
             &  &  & VOC & COCO & NUS & VG & \textit{Avg.} & VOC & COCO & NUS & VG & \textit{Avg.}\padbeforemidrule \\
            \tightmidrule \padaftermidrule
            CLIP & TaI-DPT & & 75.4 & 55.2 & \good{41.8} & 37.6 & 52.5 & 86.1 & 65.8 & 45.9 & 39.8 & 59.4 \\
            SC-CLIP & TaI-DPT & & 78.9 & 60.8 & 41.7 & 41.7 & 55.8 & 89.8 & 71.6 & 46.5 & 44.5 & 63.1 \\
            CLIP & TaI-DPT & \checkmark & 75.8 & 52.0 & 40.1 & 36.6 & 51.1 & 86.5 & 66.1 & \good{47.8} & 40.2 & 60.2 \\
            SC-CLIP & TaI-DPT & \checkmark & 78.7 & 58.6 & \best{43.8} & 41.8 & 55.7 & \good{91.4} & \good{71.9} & \best{50.6} & 45.3 & \good{64.8} \\
            SC-CLIP & LoCo & & \good{82.0} & \good{61.3} & 41.0 & \good{42.5} & \good{56.7} & 91.1 & 70.9 & 44.2 & \good{45.5} & 62.9 \\
            SC-CLIP & LoCo & \checkmark & \best{83.1} & \best{65.1} & 40.7 & \best{45.0} & \best{58.5} & \best{93.0} & \best{76.6} & 45.3 & \best{49.2} & \best{66.0} \\
            \midrule
            \multirow{2}{*}{Backbone} & \multirow{2}{*}{Method} & \multirow{2}{*}{\textit{M.S.}} & \multicolumn{5}{c}{Recall@10 (\%)} & \multicolumn{5}{c}{PR-AUC (\%)} \\
            \cmidrule(lr){4-8} \cmidrule(lr){9-13}
             & & & VOC & COCO & NUS & VG & \textit{Avg.} & VOC & COCO & NUS & VG & \textit{Avg.}\padbeforemidrule \\
            \tightmidrule \padaftermidrule
            CLIP & TaI-DPT & & 95.8 & 78.0 & 78.3 & 48.9 & 75.2 & 74.3 & 49.3 & 33.0 & 31.6 & 47.1 \\
            SC-CLIP & TaI-DPT & & 97.8 & 84.2 & 87.8 & 54.8 & 81.1 & 78.6 & 57.9 & \good{43.1} & 38.8 & 54.6 \\
            CLIP & TaI-DPT & \checkmark & 96.2 & 78.4 & 78.6 & 49.0 & 75.6 & 75.1 & 49.8 & 34.7 & 31.9 & 47.9 \\
            SC-CLIP & TaI-DPT & \checkmark & 97.7 & 84.4 & 87.6 & 54.6 & 81.1 & 78.8 & 57.5 & \best{43.2} & 39.3 & 54.7 \\
            SC-CLIP & LoCo & & \good{99.6} & \good{89.6} & \good{89.9} & \good{55.0} & \good{83.5} & \good{84.1} & \good{65.8} & 41.5 & \good{39.9} & \good{57.8} \\
            SC-CLIP & LoCo & \checkmark & \best{99.9} & \best{92.1} & \best{90.6} & \best{57.9} & \best{85.1} & \best{84.9} & \best{69.7} & 41.4 & \best{43.0} & \best{59.8} \\
            \bottomrule
        \end{NiceTabular}
    }
\end{table}

With the same SC-CLIP encoder and \textit{M.S.}, LoCo achieves higher average F1-score, mAP, Recall@10, and PR-AUC than TaI-DPT. The gains indicate that LoCo better uses the spatially consistent local features under the matched inference setting.

\FloatBarrier
\section{Compatibility Prior Analysis}
\label{sec:compatibility_prior_analysis}

\subsection{Prior Construction}
\label{sec:prior_construction}

To obtain prior knowledge about local coexistence and mutual exclusion among categories in a given label set, we query an LLM with a manually designed prompt template. For each dataset, we provide the complete category name list and ask the LLM to determine, under a semantic segmentation setting, which category concepts can simultaneously apply to the same pixel and which categories should be regarded as mutually exclusive. In our experiments, we use the GPT-5.5 Thinking model as the LLM. The prompt requests the output in the form of a Python dictionary, \texttt{dict[str, list[str]]}, where each key denotes a category name and the corresponding value lists the categories that can coexist with it at the pixel level. The LLM returns a structured dictionary-like response, in which each category is associated with a list of semantically compatible categories. We parse the responses offline into a local compatibility graph and apply OR symmetrization: a pair is compatible if either category lists the other. The resulting graph is fixed and reused for images with the same category vocabulary. Compatibility describes concepts that can apply to the same local region, rather than merely co-occurring somewhere in one image. Each category competes with all categories except itself and its compatible neighbors.
The prompt we use is given as follows.

\begin{verbatim}
In semantic segmentation scenarios, it is usually assumed that a pixel
can belong to only one concept from the category set, thus there are no
cases where multiple concepts can coexist. Now, given a category set
containing n category names, I would like to obtain a prior assumption
about which categories are mutually exclusive at the same pixel and
which category concepts can simultaneously apply to the same pixel.
Please first analyze the category set, and then generate a Python
dictionary in the form of dict[str, list[str]], listing the categories
that can coexist at the same pixel.
{{category_name_list}}
\end{verbatim}

\subsection{Generation, Prompt, and Cross-LLM Stability}
\label{sec:prior_stability}

We evaluate graph stability under three sources of variation: five independent GPT-5.5 generations, three semantically equivalent prompt variants, and four different LLMs. Each repeated generation uses a separate single-turn conversation; prompt comparisons vary the wording, and model comparisons use the same prompt. \Cref{tab:prior_generation_prompt} separates repeated generation from prompt variation. ``Shared'' counts the edges present in every graph in that condition. For edge sets $A$ and $B$, Jaccard similarity is $|A\cap B|/|A\cup B|$ and overlap is $|A\cap B|/\min(|A|,|B|)$; the tables report average pairwise similarities in percent.

\begin{table}[!htbp]
    \centering
    \setlength{\tabcolsep}{4pt}
    \caption{Compatibility-graph stability under five independent generations and three prompt variants. Edge ranges and shared-edge counts are integers; Jaccard and overlap are percentages.}
    \label{tab:prior_generation_prompt}
    {
        \tablefont
        \begin{tabular}{lrrrrrrrr}
            \toprule
            \multirow{2}{*}{Dataset} & \multicolumn{4}{c}{Five independent generations} & \multicolumn{4}{c}{Three prompt variants} \\
            \cmidrule(lr){2-5} \cmidrule(lr){6-9}
            & Edges & Shared & Jaccard & Overlap & Edges & Shared & Jaccard & Overlap \\
            \midrule
            VOC (20) & 25--42 & 25 & 77.0\% & 97.6\% & 31--45 & 26 & 63.4\% & 88.4\% \\
            COCO (80) & 213--283 & 174 & 73.1\% & 90.6\% & 190--221 & 151 & 71.8\% & 88.0\% \\
            NUS (81) & 277--428 & 221 & 69.1\% & 90.0\% & 266--367 & 210 & 63.9\% & 87.3\% \\
            VG (150) & 632--1019 & 531 & 66.7\% & 88.9\% & 593--822 & 390 & 52.3\% & 77.4\% \\
            \bottomrule
        \end{tabular}
    }
\end{table}

\Cref{tab:prior_model_overlap} reports the cross-LLM edge-set comparisons. The four-model LLM consensus is the intersection of their generated compatibility graphs; its shared-edge counts are 22, 70, 125, and 286 for VOC, COCO, NUS, and VG. Cross-model Jaccard similarity ranges from 35.8\% to 70.3\%, whereas overlap ranges from 79.7\% to 93.9\%. These summarize graph structure, separately from recognition performance.

\begin{table}[!ht]
    \centering
    \setlength{\tabcolsep}{4pt}
    \caption{Cross-LLM compatibility-graph consistency. Shared edges form the four-model LLM consensus. Similarities are reported in percent.}
    \label{tab:prior_model_overlap}
    \tablefont
    \begin{tabular}{lrrrr}
        \toprule
        Dataset & Edges & Shared & Jaccard (\%) & Overlap (\%) \\
        \midrule
        VOC (20) & 25--38 & 22 & 70.3\% & 93.9\% \\
        COCO (80) & 98--456 & 70 & 35.8\% & 88.6\% \\
        NUS (81) & 157--597 & 125 & 40.9\% & 84.5\% \\
        VG (150) & 438--882 & 286 & 48.9\% & 79.7\% \\
        \bottomrule
    \end{tabular}
\end{table}

\subsection{Recognition Sensitivity}
\label{sec:prior_recognition}

\Cref{tab:prior_sensitivity} compares recognition performance under the LLM-generated priors and their consensus. Average F1-score ranges from 58.1\% to 58.6\%, and mAP from 65.6\% to 66.1\%, despite differences in graph structure. The consensus has 125.8 edges per dataset on average, computed from the four shared-edge counts, and retains 58.6\% F1-score and 66.0\% mAP. The final row preserves the reported cross-prior standard deviation; it is not a confidence interval over repeated recognition runs.

\begin{table}[!ht]
    \centering
    \setlength{\tabcolsep}{4pt}
    \caption{Sensitivity to compatibility priors. Edges and scores are averaged over the four datasets. Scores are percentages; \best{best} and \good{second-best} values include ties at the reported precision.}
    \label{tab:prior_sensitivity}
    \tablefont
    \begin{tabular}{lrrr}
        \toprule
        Prior & \textit{Avg.} edges & \textit{Avg.} F1-score & \textit{Avg.} mAP \\
        \midrule
        glm-5.2 & 490.0 & 58.1 & \good{66.0} \\
        gemini-3.6-flash & 182.3 & 58.3 & \best{66.1} \\
        claude-opus-4.6 & 290.5 & \good{58.5} & 65.6 \\
        gpt-5.5 & 314.8 & \good{58.5} & \good{66.0} \\
        LLM consensus & 125.8 & \best{58.6} & \good{66.0} \\
        Reported std. & -- & $\pm0.2$ & $\pm0.2$ \\
        \bottomrule
    \end{tabular}
\end{table}

\subsection{WordNet-Derived Compatibility Relations}
\label{sec:wordnet_prior}

\Cref{tab:wordnet_coverage} compares the WordNet-derived graph with the LLM consensus under the current vocabularies and graph-construction rules for is-a, part-of, and synonym relations. The LLM consensus is the intersection across four LLM-generated graphs. The LLM--WordNet intersection is the further intersection of that consensus with the WordNet-derived graph; these are different sets. For the empty WordNet graph on VOC, overlap is displayed as 0.0\% by the table's reporting convention.

\begin{table}[!ht]
    \centering
    \setlength{\tabcolsep}{4pt}
    \caption{WordNet-derived relations and the four-model LLM consensus. Jaccard and overlap compare their edge sets.}
    \label{tab:wordnet_coverage}
    \tablefont
    \begin{tabular}{lrrrr}
        \toprule
        Dataset & WordNet edges & LLM-consensus edges & Jaccard (\%) & Overlap (\%) \\
        \midrule
        VOC (20) & 0 & 22 & 0.0\% & 0.0\% \\
        COCO (80) & 3 & 70 & 1.4\% & 33.3\% \\
        NUS (81) & 34 & 125 & 20.5\% & 79.4\% \\
        VG (150) & 323 & 286 & 39.7\% & 60.5\% \\
        \bottomrule
    \end{tabular}
\end{table}

On COCO, the two graphs contain 3 and 70 edges, with one edge in the LLM--WordNet intersection; 69 LLM-consensus edges are absent from the WordNet-derived graph. On NUS, the respective counts are 34 and 125, with 27 shared edges, leaving 98 LLM-consensus edges absent from the WordNet-derived graph and 7 WordNet edges outside the consensus. Reported differences include scene or functional pairs such as person--backpack and apple--bowl. These coverage observations concern the chosen construction rules and category vocabularies. The recognition comparison in \Cref{tab:compatibility_ablation} shows that the WordNet prior improves F1-score over full competition, while the LLM priors achieve higher average F1-score and mAP.

\FloatBarrier
\section{Qualitative Results and Failure Cases}
\label{sec:qualitative_results}

\begin{figure}[h]
    \centering
        \begin{minipage}[b]{0.75\linewidth}
            \centering
            \includegraphics[width=\linewidth]{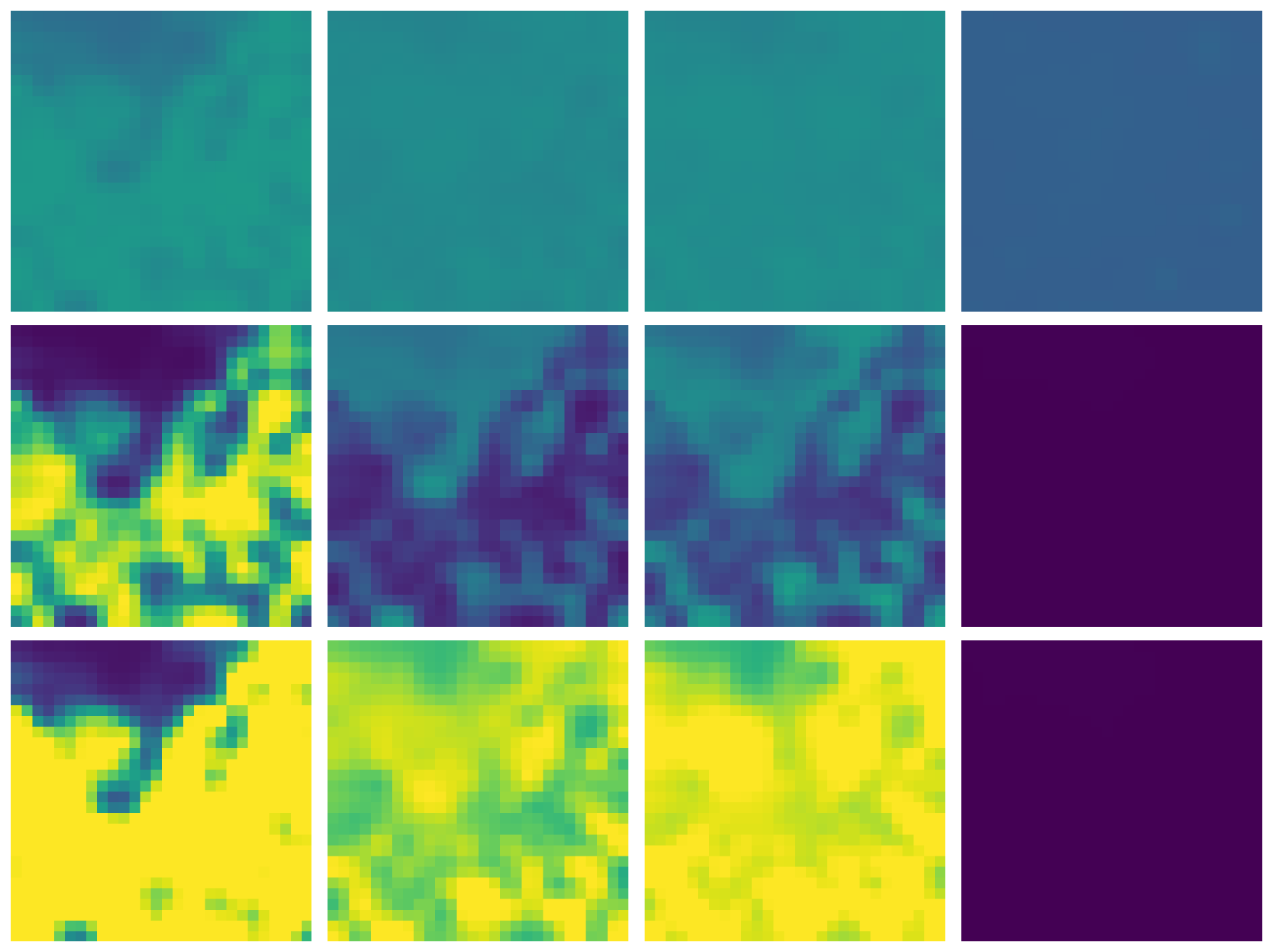}
        \end{minipage}
        \hfill
        \begin{minipage}[b]{0.18\linewidth}
            \centering
            \includegraphics[width=\linewidth]{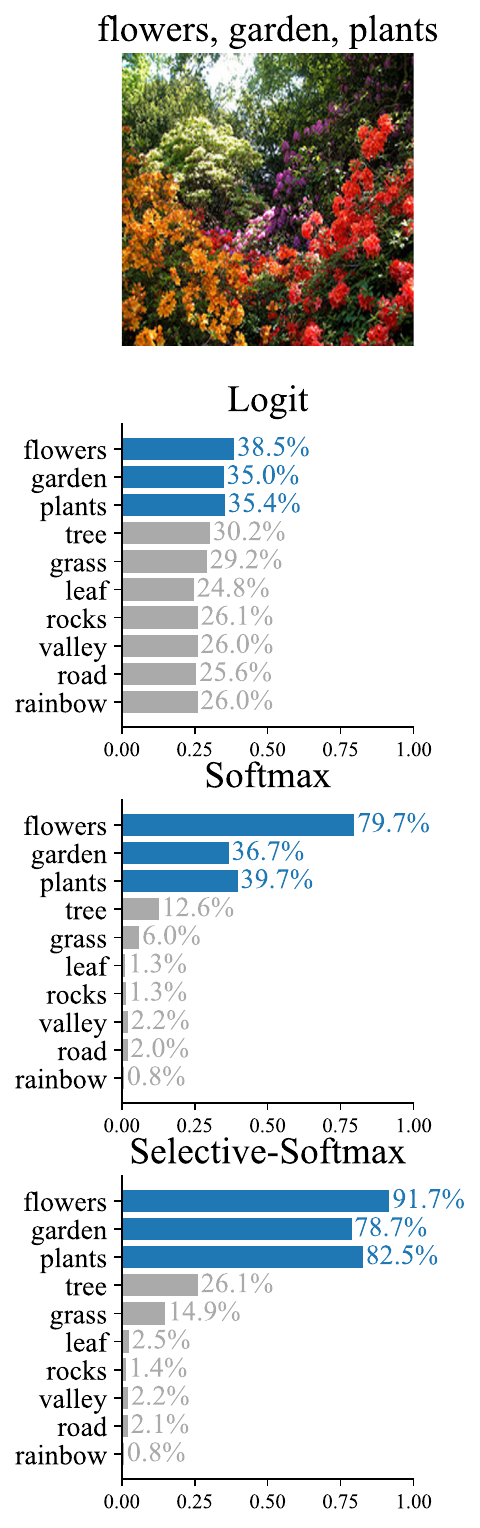}
        \end{minipage}
        \caption{More qualitative results of LoCo's dense predictions. Class activation maps in rows 1, 2, and 3 show logits, Softmax predictions, and Selective Softmax predictions, while columns 1, 2, 3, and 4 indicate \textit{flowers}, \textit{garden}, \textit{plants}, and \textit{person} classes.}
    \label{fig:cam_for_garden}
\end{figure}

\begin{figure}[t]
    \centering
        \begin{minipage}[b]{0.75\linewidth}
            \centering
            \includegraphics[width=\linewidth]{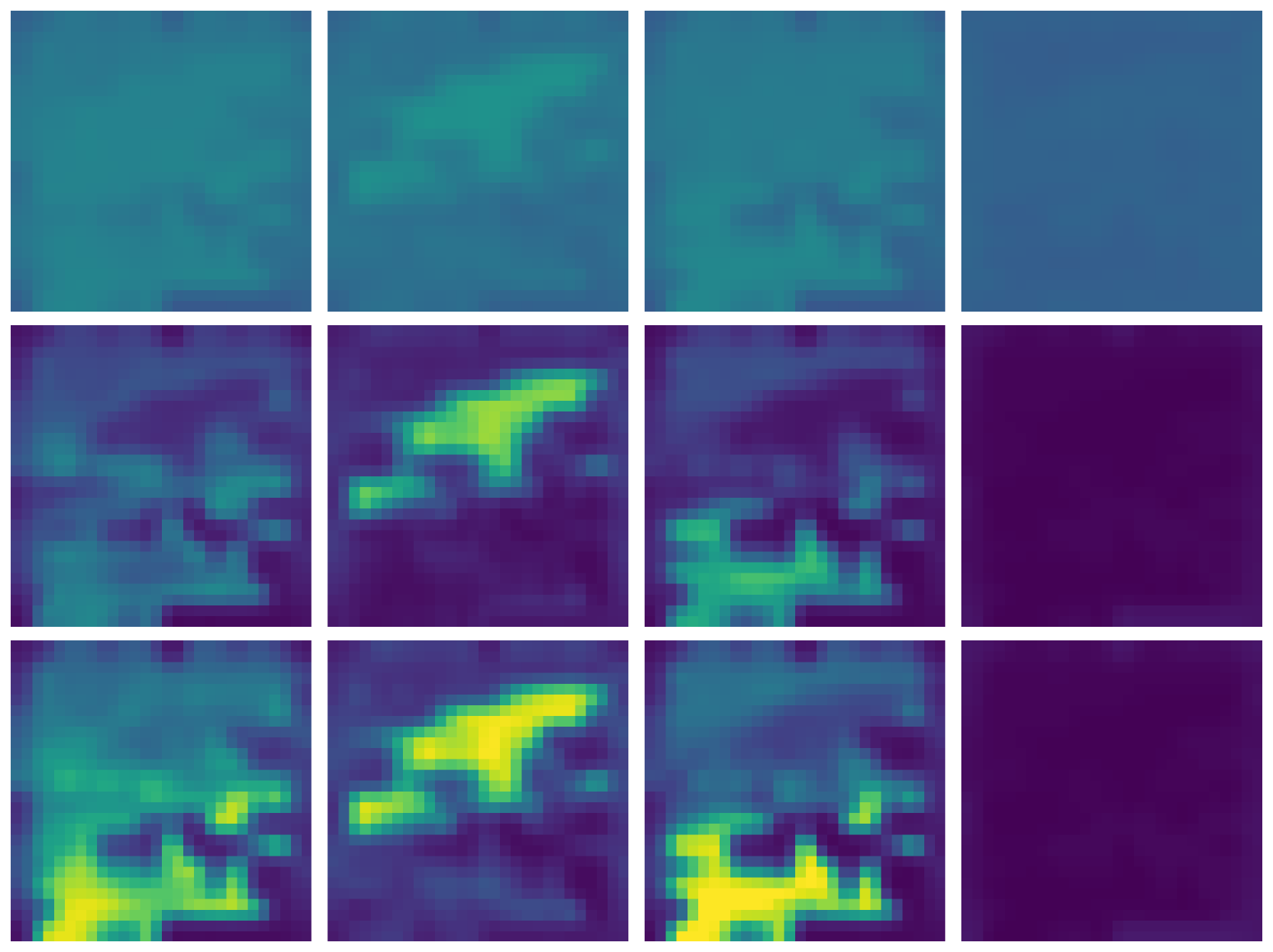}
        \end{minipage}
        \hfill
        \begin{minipage}[b]{0.18\linewidth}
            \centering
            \includegraphics[width=\linewidth]{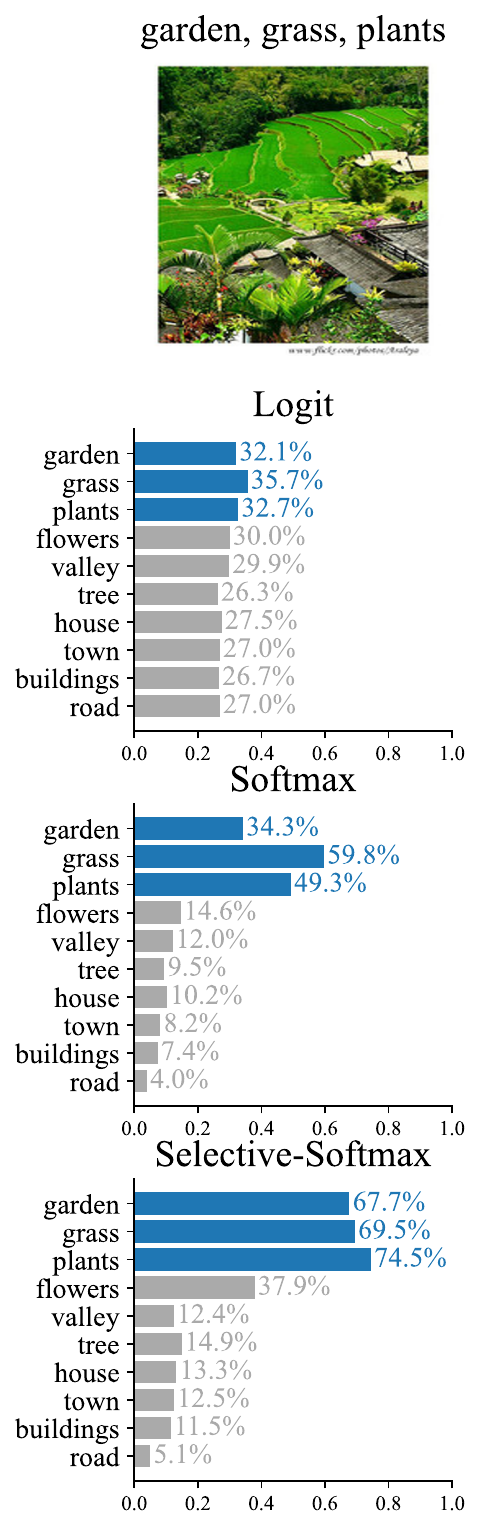}
        \end{minipage}
        \caption{More qualitative results of LoCo's dense predictions. Class activation maps in rows 1, 2, and 3 show raw logits, softmax predictions, and selective softmax predictions, while columns 1, 2, 3, and 4 indicate \textit{garden}, \textit{grass}, \textit{plants}, and \textit{person} classes.}
    \label{fig:cam_for_forest}
\end{figure}

\subsection{More qualitative results} 

Beyond the visualizations in the main paper, we consistently observe that LoCo produces more spatially concentrated and semantically cleaner local responses across different datasets, which is shown in \Cref{fig:cam_for_garden} and \Cref{fig:cam_for_forest}.
In particular, Selective Softmax allows compatible categories to remain simultaneously active on the same region, while still suppressing clearly incompatible concepts. This behavior is especially beneficial for category sets containing coarse-to-fine relations, part--whole composition, and other non-exclusive semantics, where standard softmax tends to over-suppress true positive labels.

\begin{figure}[t]
    \centering
    \begin{subfigure}[b]{0.19\linewidth}
        \centering
        \includegraphics[width=\linewidth]{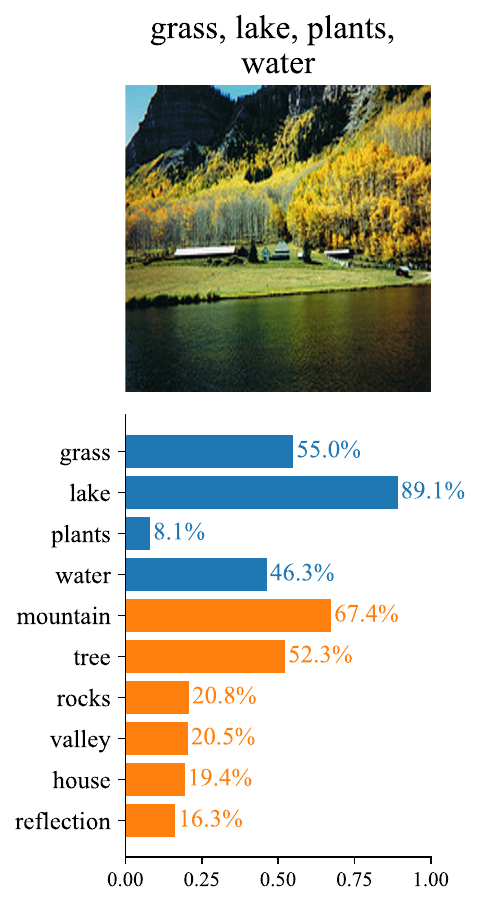}
        \subcaption{}
    \end{subfigure}
    \begin{subfigure}[b]{0.19\linewidth}
        \centering
        \includegraphics[width=\linewidth]{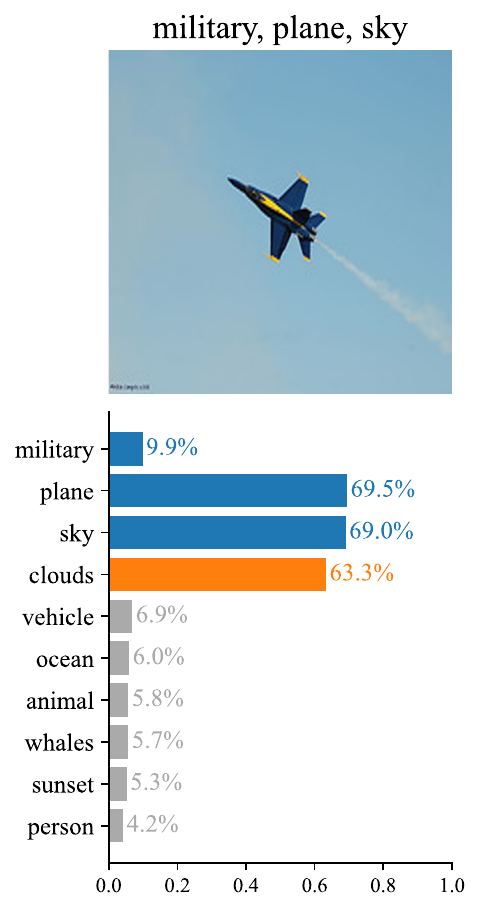}
        \subcaption{}
    \end{subfigure}
    \begin{subfigure}[b]{0.19\linewidth}
        \centering
        \includegraphics[width=\linewidth]{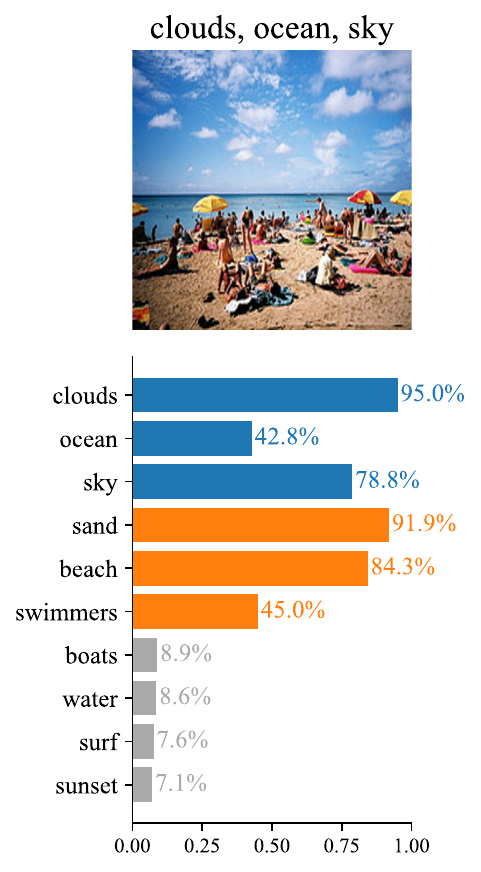}
        \subcaption{}
    \end{subfigure}
    \begin{subfigure}[b]{0.19\linewidth}
        \centering
        \includegraphics[width=\linewidth]{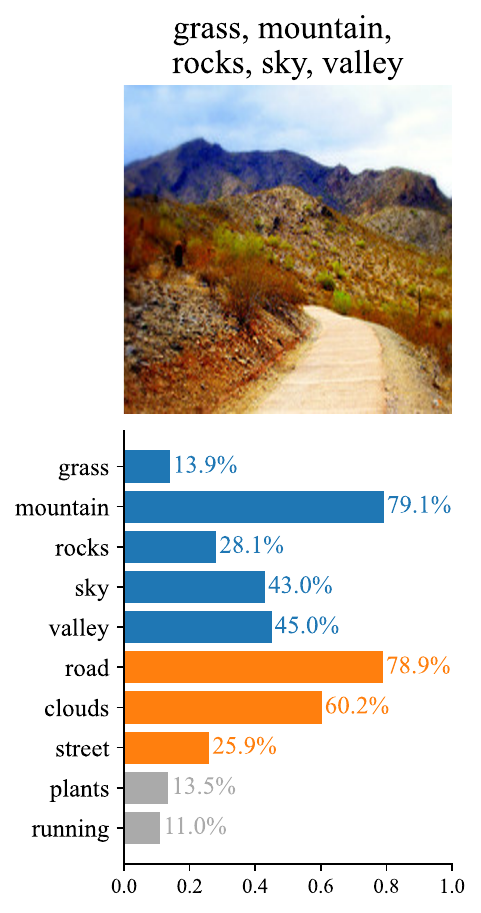}
        \subcaption{}
    \end{subfigure}
    \begin{subfigure}[b]{0.19\linewidth}
        \centering
        \includegraphics[width=\linewidth]{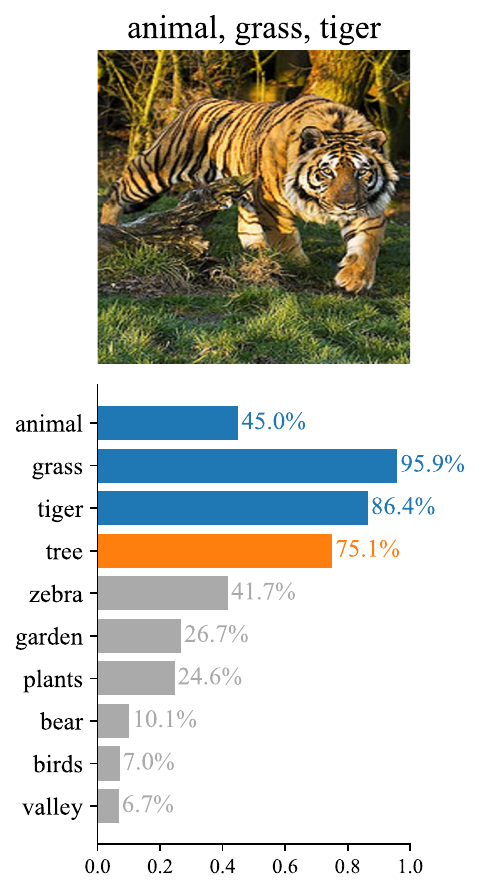}
        \subcaption{}
    \end{subfigure}
    \begin{subfigure}[b]{0.19\linewidth}
        \centering
        \includegraphics[width=\linewidth]{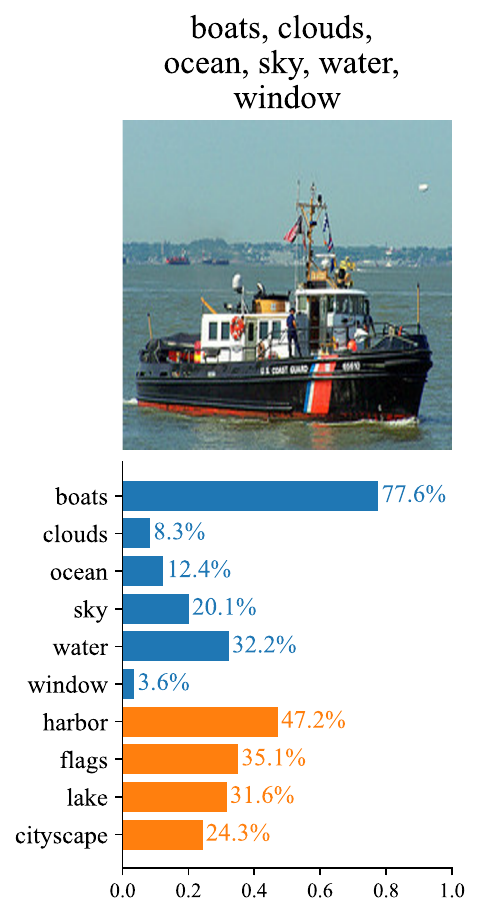}
        \subcaption{}
    \end{subfigure}
    \begin{subfigure}[b]{0.19\linewidth}
        \centering
        \includegraphics[width=\linewidth]{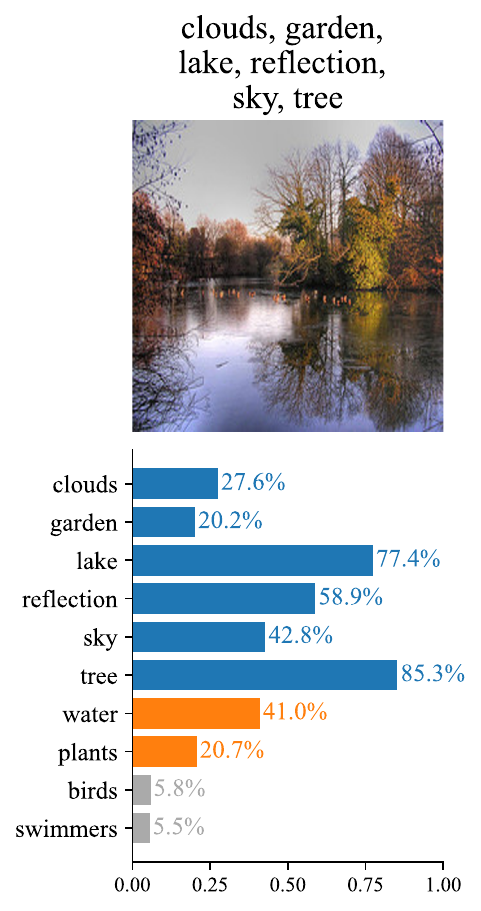}
        \subcaption{}
    \end{subfigure}
    \begin{subfigure}[b]{0.19\linewidth}
        \centering
        \includegraphics[width=\linewidth]{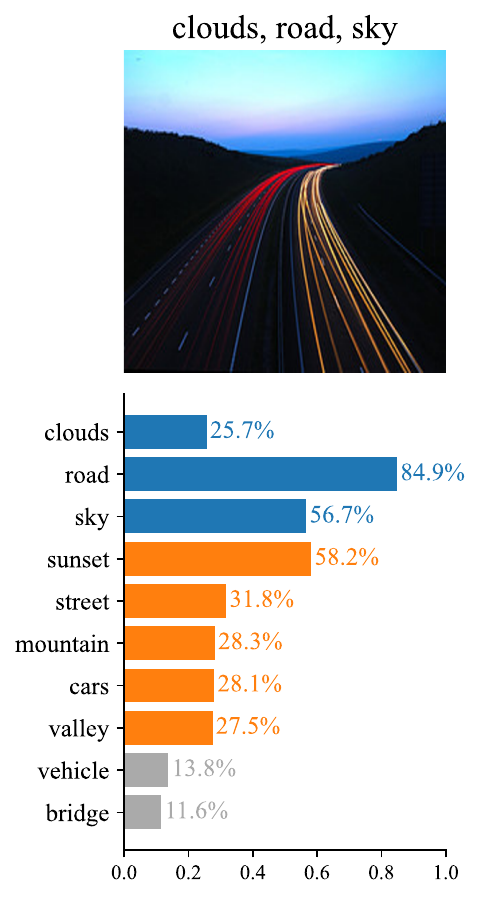}
        \subcaption{}
    \end{subfigure}
    \begin{subfigure}[b]{0.19\linewidth}
        \centering
        \includegraphics[width=\linewidth]{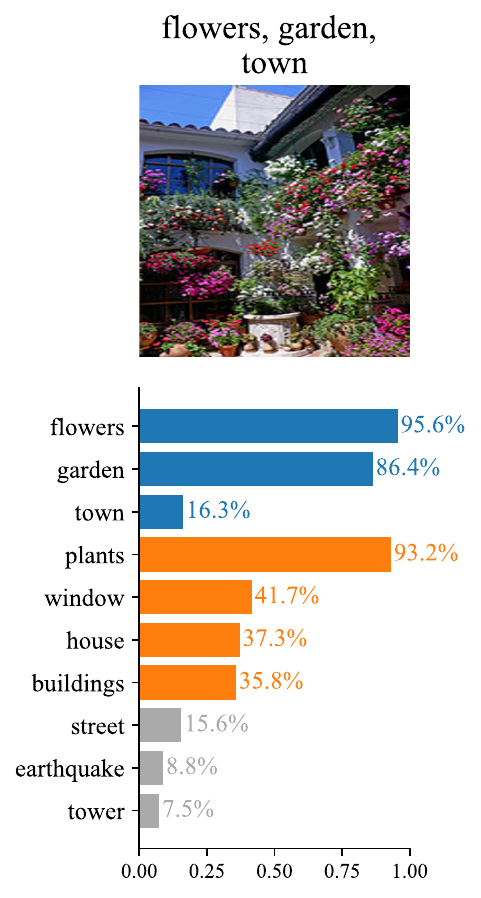}
        \subcaption{}
    \end{subfigure}
    \begin{subfigure}[b]{0.19\linewidth}
        \centering
        \includegraphics[width=\linewidth]{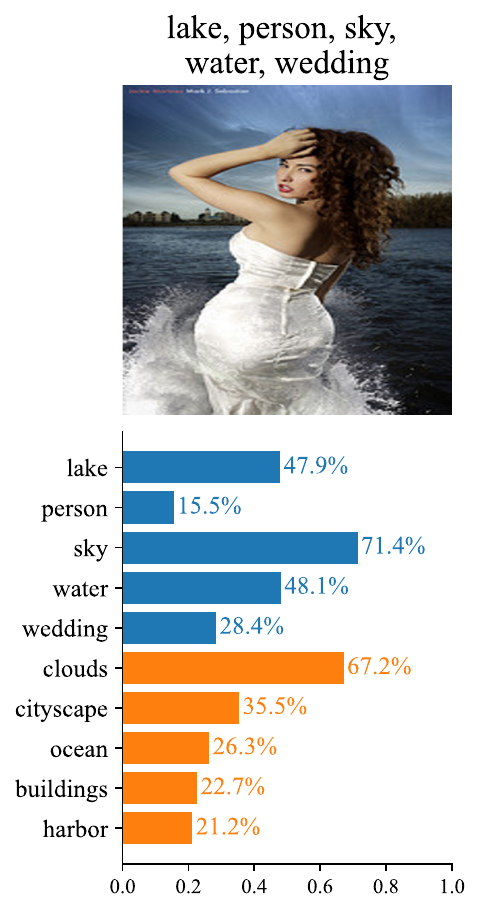}
        \subcaption{}
    \end{subfigure}
    \caption{Failure cases for LoCo on the NUS dataset. Blue bars denote scores for annotated positive categories; orange bars highlight selected categories absent from the original annotations. The original annotations are shown above each image. Some unannotated concepts are visually present.}
    \label{fig:failure_cases}
\end{figure}

\subsection{Failure cases} 

\Cref{fig:failure_cases} illustrates three qualitative challenges: small or visually weak objects, categories requiring broader scene or attribute context, and ambiguous category boundaries. Some visually present concepts are absent from the original annotations. These observations concern selected examples and inspected pairs; the benchmark comparisons retain the original annotations.

\section{Limitations and Future Work}
\label{sec:limitations}

LoCo uses a fixed binary compatibility graph for each category vocabulary. This prior approximates richer, context-dependent relations and requires compatibility relations for the target vocabulary. Context-adaptive or instance-conditioned compatibility is a direction for future work. Local multi-scale inference improves average recognition performance at additional computational cost; optional global ensembling adds a separate predictor's cost (\Cref{sec:computation_resources}).

\let\addcontentsline\appendixaddcontentsline